\documentclass[11pt]{article}
\usepackage{mhchem}

\usepackage{acl}

\usepackage{times}
\usepackage{latexsym}

\usepackage[T1]{fontenc}

\usepackage[utf8]{inputenc}

\usepackage{colortbl}
\usepackage{threeparttable}
\usepackage{pifont} 
\usepackage{multirow}
\usepackage{makecell}
\usepackage{array}
\usepackage{booktabs}
\usepackage{enumitem}
\usepackage{microtype}
\usepackage{tabularx}
\usepackage{inconsolata}

\usepackage{graphicx}

\usepackage[most]{tcolorbox}
\definecolor{blue}{HTML}{013371}
\newtcbtheorem[]{prompt}{Prompt}
{colback=gray!00,colframe=blue,fonttitle=\bfseries, left=.02in, right=.02in,bottom=.02in, top=.02in}{prompt}

\definecolor{green}{HTML}{6AA84F}
\newtcbtheorem[]{case}{Case}
{colback=gray!00,colframe=green,fonttitle=\bfseries, left=.02in, right=.02in,bottom=.02in, top=.02in}{case}

\newtcbtheorem[]{solution}{Solution}
{colback=gray!00,colframe=green,fonttitle=\bfseries, left=.02in, right=.02in,bottom=.02in, top=.02in}{solution}

\title{STEMVerse: A Dual-Axis Diagnostic Framework for STEM Reasoning in Large Language Models}

\author{\textbf{Xuzhao Li}$^{1*}$\hspace{9pt}
\textbf{Xuchen Li}$^{2}$\thanks{Equal contribution.} \hspace{9pt} 
\textbf{Jian Zhao}$^{2,3}$\hspace{9pt}
\textbf{Shiyu Hu}$^{1}$\thanks{Corresponding Author.} \hspace{9pt}\\
\textsuperscript{1}NTU,
\textsuperscript{2}ZGCA,
\textsuperscript{3}ZGCI\\
\tt\small xuzhaoli2001@gmail.com, xuchenli1030@gmail.com, shiyu.hu@ntu.edu.sg
}

\begin{document}
\maketitle
\begin{abstract}
As Large Language Models (LLMs) achieve significant breakthroughs in complex reasoning tasks, evaluating their proficiency in science, technology, engineering, and mathematics (STEM) has become a primary method for measuring machine intelligence. However, current evaluation paradigms often treat benchmarks as isolated "silos," offering only monolithic aggregate scores that neglect the intricacies of both academic specialization and cognitive depth. This result-oriented approach fails to distinguish whether model errors stem from insufficient domain knowledge or deficiencies in cognitive capacity, thereby limiting the diagnostic value. To address this, we propose STEMVerse, a diagnostic framework designed to systematically analyze the STEM reasoning capabilities of LLMs. This framework characterizes model performance across academic specialization and cognitive complexity to map the capability required for reasoning. We re-aggregate over 20,000 STEM problems from mainstream benchmarks into a unified "Discipline $\times$ Cognition" capability space, assigning dual-axis labels to every instance. Utilizing this unified diagnostic framework, we systematically evaluate representative LLM families across varying parameter scales and training paradigms. Our empirical results reveal structural failure patterns in STEM reasoning. By integrating multi-disciplinary coverage and fine-grained cognitive stratification into a unified framework, STEMVerse provides a clear and actionable perspective for understanding the scientific reasoning characteristics of LLMs.
\end{abstract}

\begin{figure}[htbp!]
    \centering
    \includegraphics[width=0.45\textwidth]{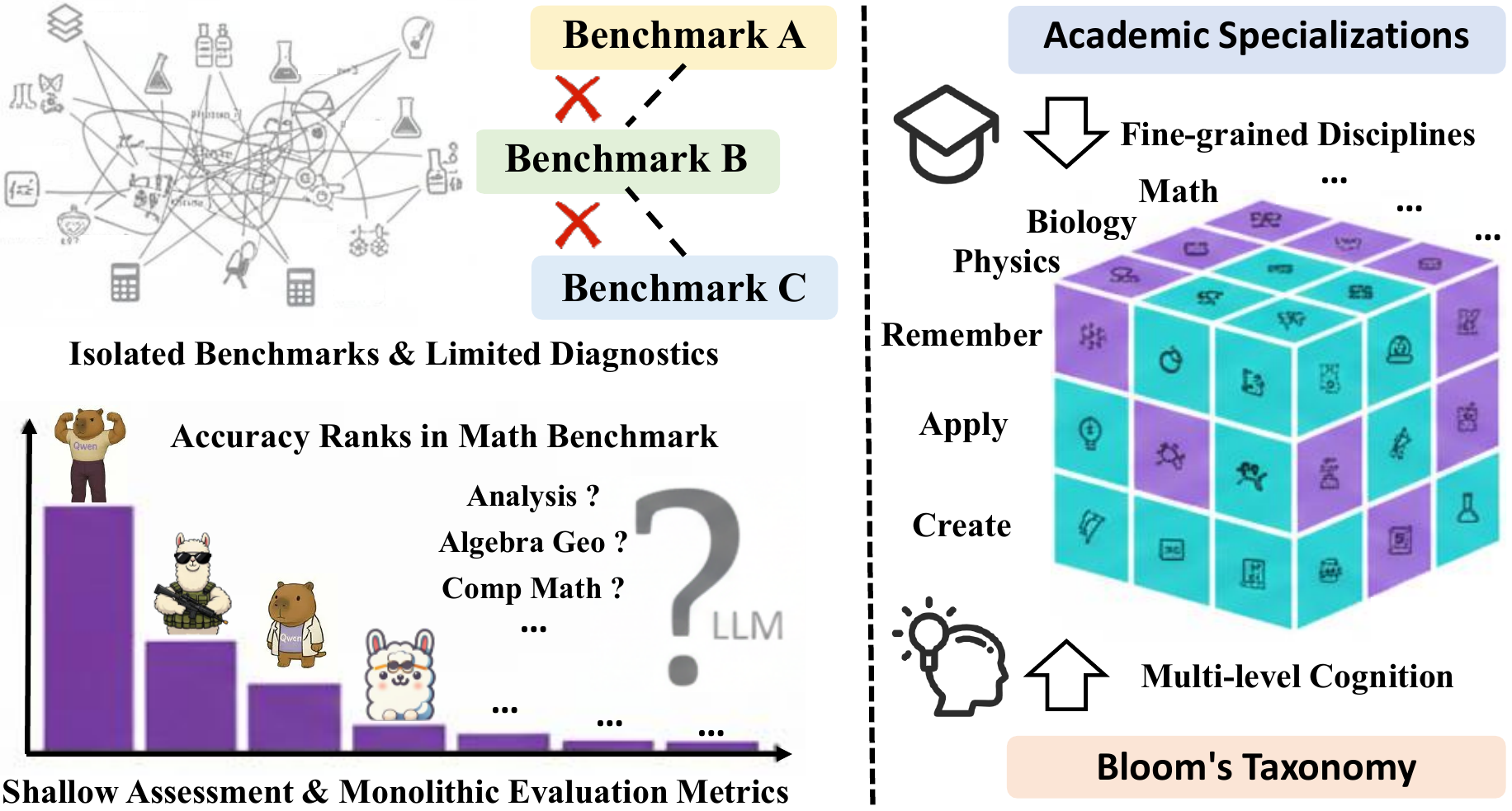}
    \vspace{-5pt}
    \caption{\textbf{Paradigm shift from result-oriented ranking to capability-driven diagnostics.} \textbf{Left:} Traditional benchmarks treat disciplines as isolated silos, offering only monolithic accuracy scores that neglect the intricacies and profundities of both academic specialization and cognitive depth. \textbf{Right:} STEMVerse restructures evaluation into a dual-axis capability matrix to pinpoint "logical blind spots."}
    \label{fig:insight}
    \vspace{-15pt}
\end{figure}

\section{Introduction}
As Large Language Models (LLMs) achieve significant breakthroughs in complex reasoning tasks \cite{guo2025deepseek,xie2025logic,guan2025rstar,gpt4o,gemini,gemini2.0flash,cao2025large}, evaluating their proficiency in science, technology, engineering and mathematics (STEM) has become a primary method for measuring machine intelligence \cite{wang2023scibench,wang2024mmlu,li2025multimodal}. The ability to solve intricate mathematical proofs, interpret physical phenomena, and analyze biological systems is no longer just a specialized requirement but a core benchmark for evaluating a model's logical rigor and internal knowledge representation. Consequently, numerous benchmarks \cite{he2024olympiadbench,huang2024olympicarena,li2024dtllm} have been developed to evaluate the scientific capabilities of LLMs.

However, as shown in Fig. \ref{fig:insight} (Left), current evaluation paradigms often treat these benchmarks \cite{shi2024math,amini2019mathqa,cobbe2021gsm8k} as isolated "silos," offering only aggregate scores \cite{bisk2020piqa,walker2010chembench} that neglect the structural intricacies of academic specialization and cognitive depth. Most existing leaderboards \cite{wang2024mmlu,li2025verifybench} report a single accuracy per benchmark, creating a "black-box" approach that fails to distinguish between different sources of model failure. For instance, a model’s failure on a complex physics problem could stem from a lack of specialized formulas (knowledge gap) or a breakdown in multi-step causal deduction (reasoning gap). Without a structured way to dissect these failures, the community lacks the necessary insights to refine model performance systematically, often mistaking task boundaries for true capability boundaries.

To bridge this gap, we propose STEMVerse, a diagnostic framework that transits LLMs evaluation from coarse-grained metrics to a "spectral" analysis of capabilities. As shown in Fig. \ref{fig:insight} (Right), STEMVerse begins by stripping away original data labels and re-aggregating discrete problems from heterogeneous benchmarks into a unified disciplinary coordinate system. At the core of our framework is a dual-axis capability matrix that meticulously projects each problem onto two orthogonal dimensions: a vertical axis representing fine-grained academic specializations (spanning 27 sub-disciplines across Mathematics, Physics, Chemistry, and Biology) and a horizontal axis employing Bloom’s Taxonomy to categorize cognitive complexity across six hierarchical levels. This structured grid allows us to pinpoint the exact intersection where a model’s reasoning fails, transforming evaluation from aggregate rankings to a principled roadmap for localizing "logical blind spots."

Utilizing STEMVerse, we conduct extensive evaluations on representative open-source model families, specifically the Qwen \cite{qwen2025qwen25technicalreport,yang2025qwen3} and Llama \cite{dubey2024llama} series, ranging from 3B to 14B parameters. Our empirical results validate the necessity of this dual-axis perspective: while aggregate scores may show steady growth, our matrix reveals a non-linear evolutionary pattern of academic and cognitive capabilities. Specifically, we identify a "logic-symbolic collapse" in symbolic-heavy fields, where models demonstrate proficiency in formulaic execution but suffer sharp performance degradation during the transition to high-order cognitive tasks. These findings underscore the limitations of current "one-score-fits-all" benchmarks and highlight the precision of STEMVerse in diagnosing the structural deficiencies of current training paradigms.

This work makes three key contributions:

\begin{itemize}[itemsep=2pt, topsep=0pt, parsep=2pt]
\item We propose STEMVerse to address the fragmented organization and result-oriented nature of STEM reasoning evaluation. By breaking traditional benchmark boundaries, we extends STEM evaluation from monolithic accuracy rankings to a comprehensive capability analysis that simultaneously accounts for disciplinary disparities and cognitive complexity.

\item We introduce an evaluation methodology based on a dual-axis capability matrix to address the difficulty in characterizing model reasoning with fine-grained precision. This approach intertwines granular academic specializations with Bloom’s Taxonomy to systematically map the distribution of model reasoning across disciplines and cognitive tiers.

\item To address the limitations of existing evaluations in revealing capability structures and evolutionary traits, we conduct systematic experiments across multiple mainstream open-source model families, parameter scales, and training paradigms. Our findings reveal cognitive bottlenecks and non-linear evolutionary patterns in STEM reasoning, providing a diagnostic foundation for understanding the inherent limitations of model capabilities.
\end{itemize}

\section{Related Work}
\subsection{Scientific Reasoning}
Scientific reasoning \cite{sciagents,ma2024sciagent} in LLMs represents a frontier in artificial intelligence, moving beyond simple pattern matching toward complex logical reasoning and symbolic manipulation \cite{narayanan2024aviary}. Recent advancements have demonstrated that models can perform sophisticated reasoning \cite{guan2025rstar,shi2024math,jaiswal2024improving,hsu2024thought,bran2025chemical} via techniques such as Chain-of-Thought (CoT) \cite{wei2022chain,zhang2022automatic}, reinforcement learning (RL) \cite{guo2025deepseek,li2025verifybench,li2025select} and multi-agent system \cite{li2025sciagent,sciagents}, which encourage step-by-step derivation. However, despite the emergence of specialized scientific models, research suggests that LLMs still struggle with multi-step causal chains and domain-specific constraints in STEM disciplines \cite{li2025system,diaz2023conceptual}. Current studies \cite{ahn2024large} focus primarily on enhancing these capabilities through fine-tuning on high-quality technical corpora or integrating external tools like calculators and code interpreters.

\begin{figure*}[htbp!]
    \vspace{-10pt}
    \centering
    \includegraphics[width=0.98\textwidth]{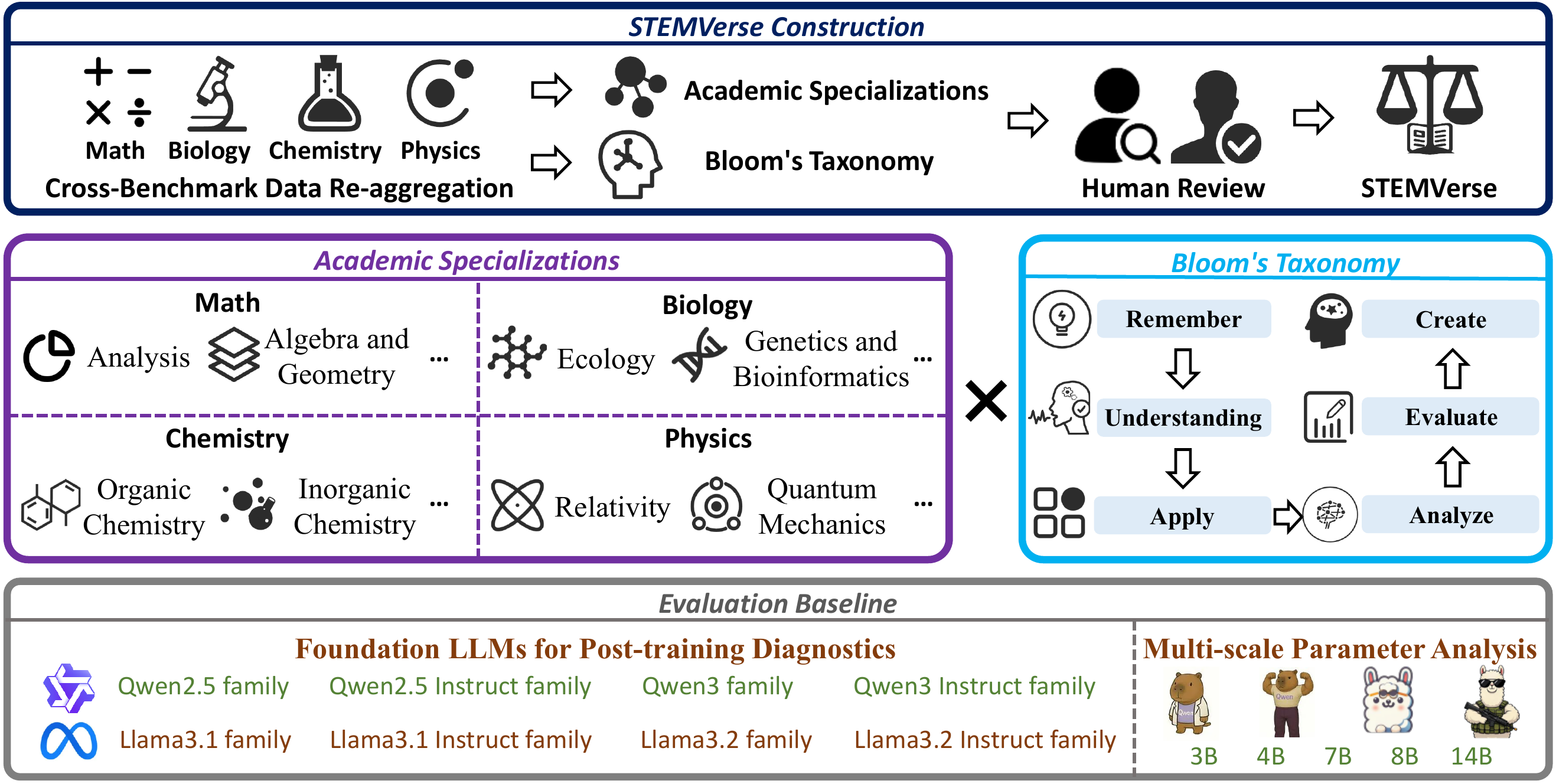}
    \vspace{-5pt}
    \caption{\textbf{Overview of the STEMVerse.} STEMVerse restructures traditional STEM benchmarks into a dual-axis capability matrix, mapping academic specializations against Bloom’s cognitive taxonomy to provide a granular, "spectral" characterization of model reasoning.}
    \label{fig:overview}
    \vspace{-15pt}
\end{figure*}

\subsection{STEM Evaluation}
As models evolve, the demand for robust evaluation frameworks has led to the development of numerous STEM-oriented benchmarks \cite{he2024olympiadbench,huang2024olympicarena}. Traditional benchmarks often categorize problems by broad subjects or rely on multiple-choice formats to track state-of-the-art performance \cite{wang2024mmlu,rein2024gpqa,du2025supergpqa}. While these benchmarks \cite{bisk2020piqa,wang2023scibench,amini2019mathqa} provide a macroscopic view of model progress, they frequently treat different benchmarks as isolated silos, offering monolithic aggregate scores \cite{walker2010chembench,cobbe2021gsm8k,li2025causalstep}. Such a "black-box" evaluation paradigm obscures the specific reasons for model failure, making it difficult to distinguish whether a model lacks specialized domain knowledge or the underlying cognitive flexibility required for scientific tasks.

\subsection{Cognitive Taxonomy}
To address the lack of granularity in performance metrics, researchers have begun exploring cognitive psychology \cite{huber2025llms} and educational theories \cite{li202511plus,hu2024fiova} to assess machine intelligence. Among these, Bloom's Taxonomy has served as a foundational framework in pedagogy for classifying learning objectives into hierarchical levels of complexity \cite{bhambri2025cognitively}. Early attempts \cite{ma2025cognitive} have been made to utilize such taxonomies to evaluate common-sense reasoning or linguistic tasks \cite{hatalis2025review,li2025look}. However, a dual-axis framework integrating fine-grained specializations and hierarchical cognitive tiers remains largely unexplored in STEM. By adopting this structural approach, we aim to provide a more "spectral" and diagnostic characterization of how scientific reasoning scales with model capacity.

\section{STEMVerse}
\subsection{Overview}
The STEMVerse transits LLMs evaluation from coarse-grained performance metrics to fine-grained academic specialization \cite{liu2025atlas} and cognitive diagnostics \cite{huber2025llms}, enabling a "spectral" analysis of model capabilities as illustrated in Fig. \ref{fig:overview}. The process begins with cross-benchmark data re-aggregation, where we break the silos of existing STEM benchmarks by stripping away original data labels and treating the collected problems as a unified corpus. At the core of the framework is the dual-axis taxonomy mapping, where each problem is meticulously projected onto two orthogonal dimensions: the academic axis, covering fine-grained academic specializations, and the cognitive axis, which employs Bloom’s taxonomy to categorize tasks across six hierarchical levels. This mapping constructs a structured grid that allows us to pinpoint the exact intersection of discipline and cognitive complexity where a model’s reasoning fails. Ultimately, this diagnostic profiling transcends mere performance ranking to provide a principled roadmap to systematically identifying and localizing "logical blind spots."

\subsection{Cross-Benchmark Data Re-aggregation}
\begin{table}[htbp]
\centering
\caption{\textbf{Statistics and Data Sources of STEMVerse.} (Coll.: College level; HS: High School level; Conc.: Conceptual level)}
\label{tab:stem_statistics}
\resizebox{0.48\textwidth}{!}{%
\begin{tabular}{lcl}
\toprule
\textbf{Discipline} & \textbf{Volume} & \textbf{Data Sources} \\
\midrule
Mathematics & 8,094 & MATH500 \cite{math} \\
 & & MathQA \cite{amini2019mathqa} \\
 & & GSM8K \cite{cobbe2021gsm8k} \\
 & & AMC \cite{He_AMC2023_2024} \\
 & & AIME (2024, 2025) \cite{Maxwell-Jia_AIME2024_2024} \\
 & & Olympiad Benchmarks \cite{he2024olympiadbench} \\
\midrule
Physics & 5,585 & MMLU (Coll./HS/Conc.) \cite{wang2024mmlu} \\
 & & PIQA \cite{bisk2020piqa} \\
 & & SciBench-Physics \cite{wang2023scibench} \\
 & & GPQA-Physics \cite{rein2024gpqa} \\
 & & Super\_GPQA-Physics \cite{du2025supergpqa} \\
\midrule
Chemistry & 5,043 & ChemBench \cite{walker2010chembench} \\
 & & MMLU (Coll./HS) \cite{wang2024mmlu} \\
 & & GPQA-Chemistry \cite{rein2024gpqa} \\
 & & Super\_GPQA-Chemistry \cite{du2025supergpqa} \\
\midrule
Biology & 1,652 & MMLU (Coll./HS) \cite{wang2024mmlu} \\
 & & GPQA-Biology \cite{rein2024gpqa} \\
 & & Super\_GPQA-Biology \cite{du2025supergpqa} \\
\midrule
\textbf{Total} & \textbf{20,374} & \\
\bottomrule
\end{tabular}%
}
\end{table}

The construction of STEMVerse begins with cross-benchmark data re-aggregation, primarily aimed at resolving the fragmentation and isolation of existing STEM evaluation benchmarks. Traditional evaluations typically treat a single benchmark as the basic unit, reporting performance within closed data distributions, which often causes task boundaries to be mistaken for capability boundaries. In contrast, the evaluation unit of STEMVerse is explicitly defined as "academic sub-discipline $\times$ cognitive complexity," requiring problems from disparate benchmarks to be realigned into a unified disciplinary coordinate system as a prerequisite for comparative analysis.

Based on this objective, we deconstruct and regroup problems from heterogeneous benchmarks into four core STEM pillars: Mathematics, Physics, Chemistry, and Biology, further subdivided into specific sub-fields. This classification is designed to characterize model reasoning emphasis and capability disparities across scientific domains rather than merely sorting topics. By mapping scattered problems into a unified disciplinary structure, we effectively mitigate the "silo effect" of traditional benchmarks and enable the systematic alignment of reasoning performance.

Within each disciplinary pillar, the selection of problems follow a consistent principle of covering diverse levels of difficulty: (1) Foundational academic knowledge, which reflects curricula or standardized knowledge to assess basic domain mastery; (2) Reasoning-intensive tasks, which emphasize reasoning processes and problem analysis to evaluate domain-specific reasoning capabilities; and (3) Cognitive challenges, which include competition-level or research-grade problems used to evaluate the model’s upper limits in complex scientific scenarios. The roles and distributions of various benchmarks within this structure are summarized in Tab. \ref{tab:stem_statistics}.

Through this cross-benchmark re-aggregation, STEMVerse establishes a robust foundation for subsequent dual-axis diagnostic analysis. This process ensures that evaluation is no longer confined to performance rankings within a single benchmark but can instead characterize performance variances across disciplines and reasoning tiers within a unified capability space, supporting fine-grained, structural analysis of LLM scientific reasoning.

\subsection{Dual-Axis Capability Matrix}
To support fine-grained diagnostic evaluation, STEMVerse introduces a dual-axis capability matrix that embeds each problem into two orthogonal dimensions: academic specialization and cognitive complexity. Unlike evaluations based solely on monolithic metrics or a single classification axis, the primary objective of this matrix is to explicitly distinguish between different sources of model failure. Specifically, it differentiates whether a performance decline stems from insufficient domain knowledge or a breakdown in high-order reasoning, thereby shifting the evaluative focus from aggregate rankings to structural capability analysis.

\subsubsection{Academic Specializations}
The vertical axis of the dual-axis matrix characterizes the degree of academic specialization involved in each problem. Along this dimension, STEMVerse further subdivides the four core natural science pillars into a comprehensive set of academic sub-disciplines \cite{liu2025atlas}, designed to assess the model's knowledge depth and reasoning emphasis across different scientific directions. Rather than serving as mere topical tags, this disciplinary classification is intended to characterize systemic capability variances that models may exhibit across different academic domains. By aligning problems with specific sub-disciplines and incorporating an "Others" category to ensure exhaustive coverage, STEMVerse can determine whether performance bottlenecks stem from a lack of specialized knowledge in specific disciplinary directions. The complete disciplinary hierarchy and sub-discipline definitions are summarized in Tab. \ref{tab:sub_discipline_taxonomy}.

\begin{table}[htbp]
\centering
\caption{\textbf{Taxonomy of Academic Sub-disciplines.}}
\vspace{-10pt}
\label{tab:sub_discipline_taxonomy}
\resizebox{0.48\textwidth}{!}{%
\begin{tabular}{lp{8cm}}
\toprule
\textbf{Core Pillar} & \textbf{Sub-disciplines} \\
\midrule
Mathematics & Analysis; Statistics and Operations Research; Algebra and Geometry; Differential Equations and Dynamical Systems; Computational Mathematics; Interdisciplinary Mathematics. \\
\midrule
Physics     & Relativity; Astrophysics; Thermodynamics and Statistical Physics; Electrodynamics; Quantum Mechanics; Classical Mechanics; Fluid Mechanics. \\
\midrule
Chemistry   & Physical Chemistry; Inorganic Chemistry; Organic Chemistry; Analytical Chemistry; Chemical Engineering and Technology; Theoretical and Computational Chemistry. \\
\midrule
Biology     & Molecular Biology and Biotechnology; Genetics and Bioinformatics; Immunology; Physiology and Integrative Biology; Neuroscience and Psychology; Ecology; Biophysics and Biochemistry; Cell Biology. \\
\bottomrule
\end{tabular}%
}
\vspace{-15pt}
\end{table}

\subsubsection{Cognitive Complexity}
The horizontal axis of the dual-axis matrix characterizes the cognitive complexity required to solve a given problem. Drawing upon Bloom’s Taxonomy \cite{huber2025llms}, we categorize problems into six hierarchical cognitive levels, ranging from foundational knowledge retrieval and comprehension to high-order cognitive activities involving synthesis, evaluation, and creative reasoning. This dimension is designed to characterize performance variances across reasoning depths, rather than merely assessing the mastery of knowledge.

By intersecting the academic specialization dimension with the cognitive complexity dimension, STEMVerse maps the distribution of model reasoning behaviors within a unified capability space, forming a structured capability spectrum. This dual-axis diagnostic approach ensures that evaluation transcends aggregate accuracy; it enables a precise distinction between failures rooted in insufficient domain knowledge and those caused by a breakdown in high-order reasoning chains, thereby providing a clear diagnostic lens for subsequent experimental analysis.

\subsection{Annotation and Human Review}
To translate the raw data into our dual-axis matrix, we implement a hybrid annotation pipeline that leverages the efficiency of LLMs alongside the rigorous precision of human experts. This process ensures that each problem is assigned an accurate disciplinary and cognitive label.

\subsubsection{AI-Assisted Annotation}
We employ GPT-4o \cite{gpt4o} as our primary annotator to categorize the re-aggregated benchmarks. By utilizing carefully system prompts (see Appendix \ref{appendix:prompt}), the model analyzes each question and its corresponding answer to determine: (1) the specific academic specialization from our predefined taxonomy, and (2) the appropriate Bloom’s cognitive Level. This automated phase establishes a consistent baseline for our capability matrix.

\begin{figure}[t!]
    \centering
    \includegraphics[width=0.48\textwidth]{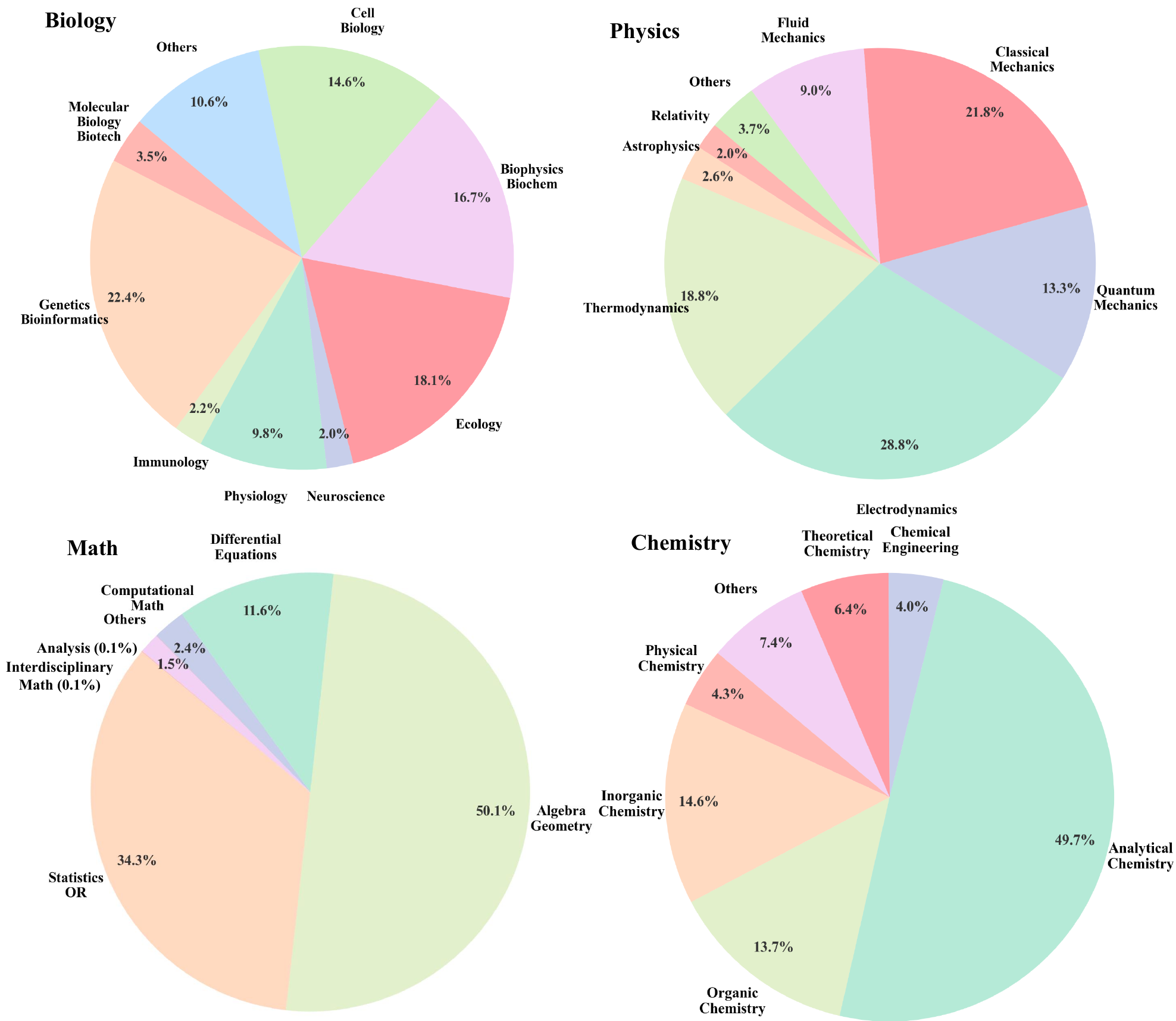}
    \vspace{-15pt}
    \caption{\textbf{Distribution of academic specializations.} The composition of fine-grained academic specializations across Mathematics, Physics, Chemistry, and Biology, ensures a balanced and comprehensive coverage of the STEM knowledge landscape.}
    \label{fig:Statistics_field}
    \vspace{-10pt}
\end{figure}

\begin{figure}[t!]
    \centering
    \includegraphics[width=0.48\textwidth]{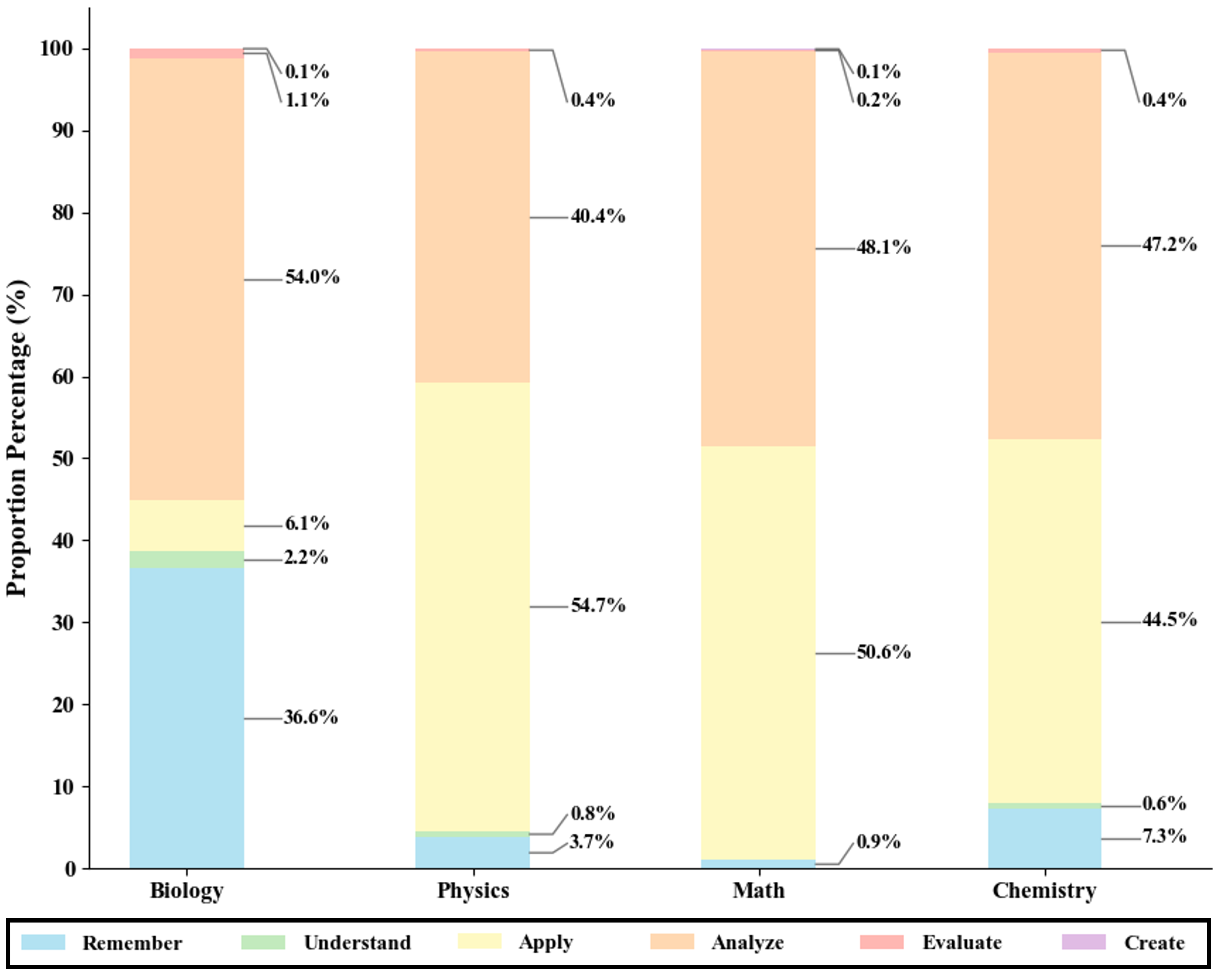}
    \vspace{-15pt}
    \caption{\textbf{Distribution of cognitive levels across disciplines.} The stacked bar chart shows the percentage of problems categorized under each level of Bloom’s Taxonomy for Biology, Physics, Mathematics, and Chemistry, highlighting the benchmark's focus on high-order reasoning evaluation.}
    \label{fig:Statistics_bloom}
    \vspace{-20pt}
\end{figure}

\begin{figure*}[t!]
    \centering
    \includegraphics[width=\textwidth]{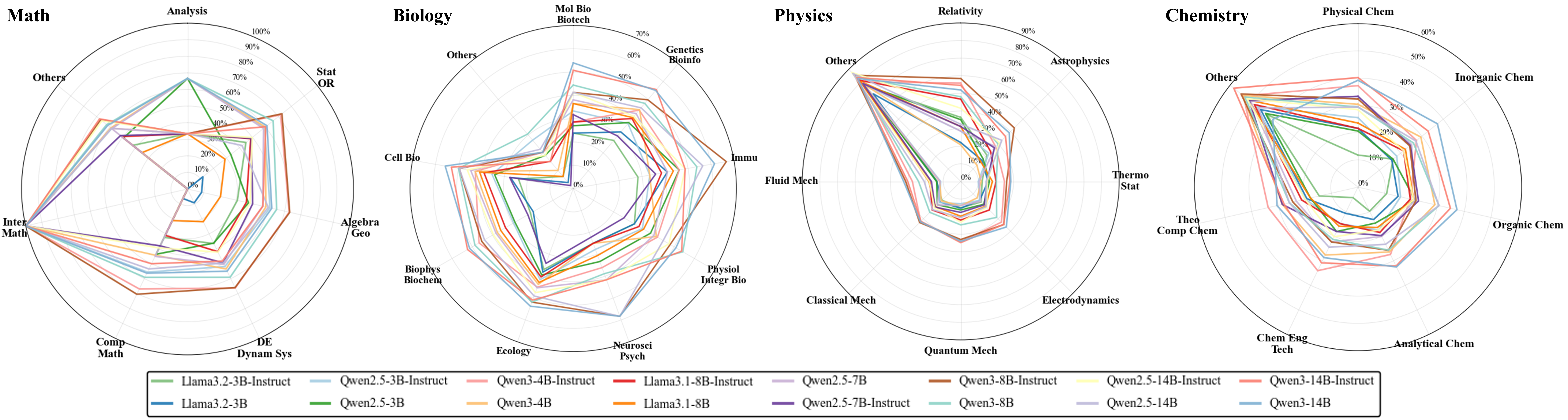}
    \vspace{-15pt}
    \caption{\textbf{Performance across academic specializations.} These radar charts visualize the accuracy of the Qwen and Llama families across fine-grained specializations in Math, Biology, Physics, and Chemistry, illustrating the performance variances. For detailed results, please refer to Appendix \ref{appendix:academic}.}
    \label{fig:results_field}
    \vspace{-15pt}
\end{figure*}

\subsubsection{Expert Manual Review and Validation}
To ensure the scientific reliability of the automated labeling, we introduced a rigorous expert manual audit process. We randomly sampled 10\% of the problems from each discipline for manual review, conducted by Master's and PhD students with the relevant academic backgrounds. The audit focused on evaluating: (1) whether the assigned academic sub-disciplines accurately reflect the core knowledge content of the problems; and (2) whether the annotated Bloom’s cognitive levels reasonably characterize the cognitive complexity required for problem-solving. For samples where discrepancies occur, a second verification is conducted by an additional expert to minimize subjective bias.

This hybrid approach effectively mitigates the "black-box" risks associated with automatic labeling by combining the scalability of GPT-4o with the nuanced judgment of domain experts. The reliability of this process is quantified through Inter-Annotator Agreement (IAA) scores. Across the four core disciplines, the IAA scores range from 0.87 to 0.92. Ultimately, this rigorous validation produces a reliable, high-fidelity benchmark that serves as a solid foundation for the subsequent evaluation and diagnostic analysis of various LLMs.

\subsection{Statistics}
The STEMVerse comprises a diverse corpus of 20,374 high-quality problems across four foundational disciplines. The benchmark is strategically distributed with Mathematics (39.7\%) and Physics (27.4\%) forming the core analytical pillars, while Chemistry (24.8\%) and Biology (8.1\%) provide specialized domain-specific challenges.

To ensure the diagnostic utility of the framework, we provide a detailed statistical analysis of the benchmark across the proposed dual-axis capability matrix. The disciplinary distribution within each pillar, illustrated in Fig. \ref{fig:Statistics_field}, is comprehensive and covers both fundamental and specialized topics. The distribution of Bloom’s cognitive levels, as detailed in Fig. \ref{fig:Statistics_bloom}, reveals the depth of reasoning required by the STEMVerse, where Analyze and Apply constitute the most significant proportions across all disciplines.

\begin{figure*}[t!]
    \centering
    \includegraphics[width=0.73\textwidth]{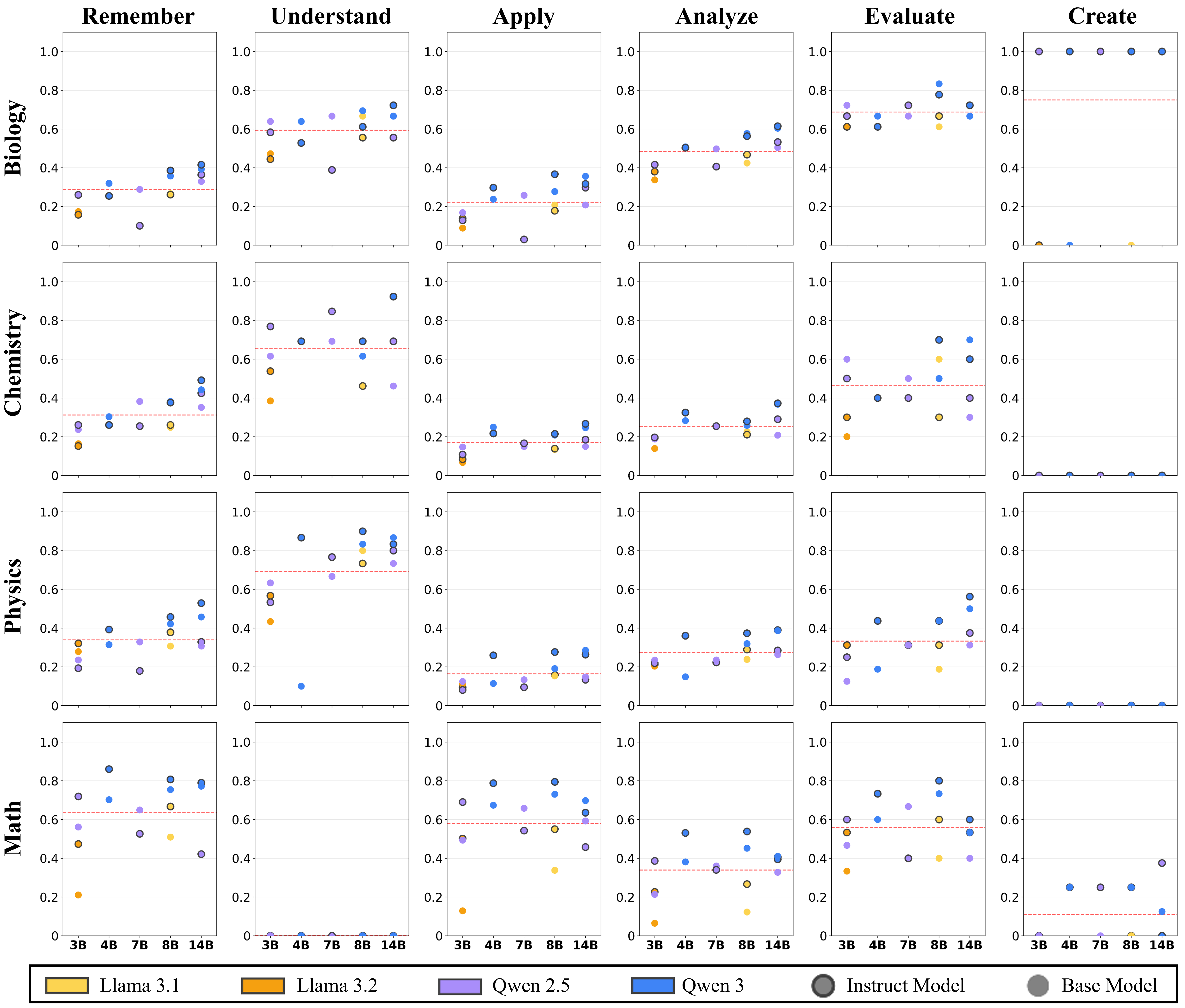}
    \vspace{-5pt}
    \caption{\textbf{Fine-grained performance diagnostics across the Dual-Axis Capability Matrix.} The matrix maps model accuracy across academic specializations and hierarchical cognitive tiers. Bubbles represent individual models from the Qwen and Llama families, categorized by parameter scale and alignment stage.}
    \label{fig:results_bloom2}
    \vspace{-15pt}
\end{figure*}

\section{Experiment}
\subsection{Baselines}
To systematically analyze the evolutionary characteristics of LLMs in STEM reasoning and to support capability comparisons within our dual-axis diagnostic framework, we selected a representative set of open-source models from the Qwen \cite{qwen2025qwen25technicalreport,yang2025qwen3} and Llama \cite{dubey2024llama} families as evaluation baselines. The core consideration for model selection was not a simple performance ranking, but rather achieving coverage across diverse parameter scales and training paradigms. This allows us to observe how model capabilities shift as disciplinary depth and cognitive complexity progressively increase. Specifically, the selected models include both base models and instruction-tuned models to distinguish the distinct roles of capacity expansion versus alignment training in STEM reasoning. Furthermore, the selection spans a wide range of parameter scales, from 3B to 14B, to characterize the evolutionary behavior of capabilities as model size scales. By evaluating these models within the same dual-axis capability space, we can perform a unified comparison of capability distributions across different model families and scale configurations, establishing a consistent reference foundation for subsequent analysis.

\subsection{Evaluation Protocol}
The evaluation protocol is aligned with the dual-axis capability matrix, aiming to project performance onto a unified space of academic specialization and cognitive complexity. Departing from the traditional approach of reporting only aggregate scores, we calculate model performance across each academic sub-discipline and Bloom’s cognitive tier, capturing fine-grained diagnostic signals.

Regarding the reasoning setup, we adopt a few-shot prompting strategy consistent with MegaScience \cite{fan2025megascience} to minimize the impact of prompting disparities on cross-model comparability. The primary evaluation metric is Accuracy, used to measure the problem-solving success rate within specific disciplinary and cognitive dimensions. It is important to emphasize that this accuracy is not intended as a standalone ranking criterion, but rather as a localized observable within the dual-axis capability matrix. By aggregating these localized accuracy results across the entire dual-axis matrix, we construct the capability spectrum of each model. This allows for a systematic analysis of reasoning patterns across different architectures and parameter scales across various disciplines and cognitive complexity.

\begin{figure*}[t!]
    \centering
    \includegraphics[width=\textwidth]{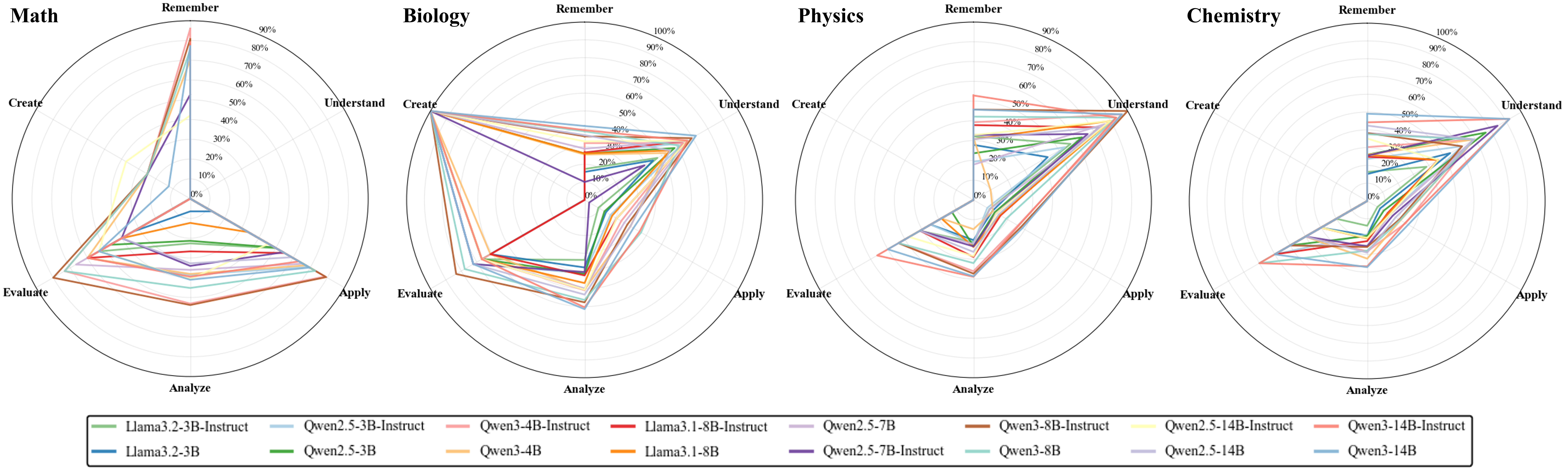}
    \vspace{-10pt}
    \caption{\textbf{Performance across cognitive levels.} These radar charts illustrate model accuracy across the six Bloom’s Taxonomy tiers for Math, Biology, Physics, and Chemistry, highlighting a general performance decrease as tasks move from Remember to Create. For detailed results, please refer to Appendix \ref{appendix:cognitive}.}
    \label{fig:results_bloom}
    \vspace{-15pt}
\end{figure*}

\section{Main Results and Analysis}
\subsection{Disciplinary Specializations}
Across the fine-grained academic specializations, different models exhibit a clear performance hierarchy (Fig. \ref{fig:results_field}). Qwen3-14B-Instruct maintains a dominant position in the vast majority of disciplines, achieving a 32.5\% accuracy in Analytical Chemistry and 58.3\% in Neuroscience and Psychology. In contrast, the Llama3.2-3B series shows significant numerical fluctuations, particularly in biology, where its accuracy in Classical Mechanics (16.7\%) is substantially lower than in Genetics \& Bioinformatics (27.0\%). Furthermore, no model below 14B parameters managed to surpass the 38.0\% accuracy threshold in Physical Chemistry.

These results reveal a "siloed knowledge" effect in smaller models. The cross-family performance inversion, where Qwen2.5-7B (25.1\%) outperforms the larger Llama3.1-8B (21.39\%) in Inorganic Chemistry, suggests that data composition during pre-training is a more reliable predictor of STEM success than raw parameter count. Qwen’s performance indicates a higher density of high-quality scientific tokens, providing a more stable foundation for specialized sub-disciplines that remains resilient across different model scales.

\subsection{Cognitive Complexity}
Model performance does not follow a simple linear decline as Bloom’s Taxonomy levels ascend (Fig. \ref{fig:results_bloom}). Performance consistently peaks at the Understand level, significantly outperforming other dimensions. However, a noticeable "performance dip" occurs upon entering the Apply stage for Biology, Physics, and Chemistry. Mathematics exhibits a unique trajectory: models maintain high proficiency in Apply tasks (e.g., Qwen2.5-7B-Instruct at 54.3\%) but suffer a sharp collapse at the Analyze stage (dropping to 34.0\%). In the highest-order Evaluate and Create dimensions, scores are extremely sparse, with Llama3.1-8B recording 0\% in Physics for Create tasks.

This reveals a logic-symbolic collapse in symbolic-heavy fields. While models excel at formulaic execution (Apply), they fail during the transition to Analyze, where tasks shift from rule-following to the decomposition of multi-stage logical chains. This divergence identifies a structural gap: while domain-specific information is successfully internalized, its reliable deployment within cognitive frameworks remains inconsistent. Models appear to possess scientific facts but lack the structural reasoning integrity required to maintain coherence as cognitive complexity increases.

\subsection{Scaling, Robustness and Training Effects}
The relationship between parameter and performance is non-linear. In the Remember tier, the Qwen3 family exhibits a predictable, incremental scaling pattern (approx. +10\% per scale jump), suggesting that increasing parameter density directly expands the model's internal "scientific database." However, the Understand tier follows a non-linear threshold; for instance, scaling from 8B to 14B triggers a massive leap from 60\% to 90\%, whereas the jump from 4B to 8B yields negligible gains. This suggests that mastering relational scientific knowledge requires a minimum parameter size to synthesize disparate concepts into a coherent framework successfully.

Furthermore, we identify an "Instruction-Tuning Paradox": the Qwen3-14B Base model consistently outperforms its Instruct counterpart in specialized Mathematics sub-disciplines (66.7\% vs 33.3\% in Analysis). While Instruction Tuning (IT) enhances format adherence and controllability, it may inadvertently suppress the diverse internal reasoning paths activated during pre-training, leading to a degradation of complex logic. These findings suggest that current training paradigms achieve horizontal expansion (more facts) at the expense of vertical reasoning depth, highlighting a structural deficiency in how alignment affects high-order reasoning. We provide some cases in Appendix \ref{appendix:case}.

\section{Conclusion}
We propose STEMVerse, a dual-axis diagnostic framework that unifies academic specialization with Bloom’s Taxonomy. By systematically characterizing the distribution of STEM reasoning capabilities, we overcome the limitations of traditional benchmarks that rely on monolithic accuracy to evaluate models. Experimental results demonstrate that while current mainstream LLMs perform reliably in knowledge retrieval and low-order cognitive tasks, they suffer from significant performance degradation at higher-order cognitive levels. This trend remains consistent across different disciplines and model scales, revealing a fundamental disconnect between the expansion of model capacity and the refinement of reasoning structures. Further analysis indicates that while alignment and instruction-tuning enhance controllability, they may inadvertently weaken a model’s symbolic reasoning and multi-step logical capabilities. This highlights a structural deficiency in existing training paradigms regarding high-order scientific reasoning.

\clearpage
\section*{Limitations}
In this study, we primarily focus on the foundational pillars of STEM: Mathematics, Physics, Chemistry, and Biology. We prioritize these subjects because they provide the most rigorous and formal logical frameworks necessary for evaluating high-order scientific reasoning. These disciplines possess well-defined symbolic systems and clear causal structures, which are essential for a principled diagnostic using Bloom’s Taxonomy. However, the current version of STEMVerse does not yet encompass more applied fields. Evaluating LLMs in these areas often requires assessing multi-modal understanding or code-execution capabilities, which are beyond the current scope of our text-based reasoning matrix. We plan to extend our dual-axis framework to these applied STEM sectors in subsequent updates.

\bibliography{custom}

@article{sciagents,
  title={SciAgents: automating scientific discovery through bioinspired multi-agent intelligent graph reasoning},
  author={Ghafarollahi, Alireza and Buehler, Markus J},
  journal={Advanced Materials},
  volume={37},
  number={22},
  pages={2413523},
  year={2025},
  publisher={Wiley Online Library}
}

@article{ma2024sciagent,
  title={Sciagent: Tool-augmented language models for scientific reasoning},
  author={Ma, Yubo and Gou, Zhibin and Hao, Junheng and Xu, Ruochen and Wang, Shuohang and Pan, Liangming and Yang, Yujiu and Cao, Yixin and Sun, Aixin and Awadalla, Hany and others},
  journal={arXiv preprint arXiv:2402.11451},
  year={2024}
}

@article{narayanan2024aviary,
  title={Aviary: training language agents on challenging scientific tasks},
  author={Narayanan, Siddharth and Braza, James D and Griffiths, Ryan-Rhys and Ponnapati, Manu and Bou, Albert and Laurent, Jon and Kabeli, Ori and Wellawatte, Geemi and Cox, Sam and Rodriques, Samuel G and others},
  journal={arXiv preprint arXiv:2412.21154},
  year={2024}
}

@article{guan2025rstar,
  title={rStar-Math: Small LLMs Can Master Math Reasoning with Self-Evolved Deep Thinking},
  author={Guan, Xinyu and Zhang, Li Lyna and Liu, Yifei and Shang, Ning and Sun, Youran and Zhu, Yi and Yang, Fan and Yang, Mao},
  journal={arXiv preprint arXiv:2501.04519},
  year={2025}
}

@inproceedings{shi2024math,
  title={Math-llava: Bootstrapping mathematical reasoning for multimodal large language models},
  author={Shi, Wenhao and Hu, Zhiqiang and Bin, Yi and Liu, Junhua and Yang, Yang and Ng, See Kiong and Bing, Lidong and Lee, Roy Ka-Wei},
  booktitle={Findings of the Association for Computational Linguistics: EMNLP 2024},
  pages={4663--4680},
  year={2024}
}

@article{jaiswal2024improving,
  title={Improving physics reasoning in large language models using mixture of refinement agents},
  author={Jaiswal, Raj and Jain, Dhruv and Popat, Harsh Parimal and Anand, Avinash and Dharmadhikari, Abhishek and Marathe, Atharva and Shah, Rajiv Ratn},
  journal={arXiv preprint arXiv:2412.00821},
  year={2024}
}

@inproceedings{hsu2024thought,
  title={Thought graph: Generating thought process for biological reasoning},
  author={Hsu, Chi-Yang and Cox, Kyle and Xu, Jiawei and Tan, Zhen and Zhai, Tianhua and Hu, Mengzhou and Pratt, Dexter and Chen, Tianlong and Hu, Ziniu and Ding, Ying},
  booktitle={Companion Proceedings of the ACM Web Conference 2024},
  pages={537--540},
  year={2024}
}

@article{bran2025chemical,
  title={Chemical reasoning in LLMs unlocks steerable synthesis planning and reaction mechanism elucidation},
  author={Bran, Andres M and Neukomm, Theo A and Armstrong, Daniel P and Jon{\v{c}}ev, Zlatko and Schwaller, Philippe},
  journal={arXiv preprint arXiv:2503.08537},
  year={2025}
}

@article{wei2022chain,
  title={Chain-of-thought prompting elicits reasoning in large language models},
  author={Wei, Jason and Wang, Xuezhi and Schuurmans, Dale and Bosma, Maarten and Xia, Fei and Chi, Ed and Le, Quoc V and Zhou, Denny and others},
  journal={Advances in neural information processing systems},
  volume={35},
  pages={24824--24837},
  year={2022}
}

@inproceedings{zhang2022automatic,
  title={Automatic chain of thought prompting in large language models},
  author={Zhang, Zhuosheng and Zhang, Aston and Li, Mu and Smola, Alex},
  booktitle={The eleventh international conference on learning representations},
  year={2022}
}

@article{guo2025deepseek,
  title={Deepseek-r1: Incentivizing reasoning capability in llms via reinforcement learning},
  author={Guo, Daya and Yang, Dejian and Zhang, Haowei and Song, Junxiao and Zhang, Ruoyu and Xu, Runxin and Zhu, Qihao and Ma, Shirong and Wang, Peiyi and Bi, Xiao and others},
  journal={arXiv preprint arXiv:2501.12948},
  year={2025}
}

@article{xie2025logic,
  title={Logic-rl: Unleashing llm reasoning with rule-based reinforcement learning},
  author={Xie, Tian and Gao, Zitian and Ren, Qingnan and Luo, Haoming and Hong, Yuqian and Dai, Bryan and Zhou, Joey and Qiu, Kai and Wu, Zhirong and Luo, Chong},
  journal={arXiv preprint arXiv:2502.14768},
  year={2025}
}

@article{li2025sciagent,
  title={SciAgent: A Unified Multi-Agent System for Generalistic Scientific Reasoning},
  author={Li, Xuchen and Wu, Ruitao and Liu, Xuanbo and Wang, Xukai and Hu, Jinbo and Bai, Zhixin and Zeng, Bohan and Liang, Hao and Chen, Leheng and Chen, Mingrui and others},
  journal={arXiv preprint arXiv:2511.08151},
  year={2025}
}

@article{li2025system,
  title={From system 1 to system 2: A survey of reasoning large language models},
  author={Li, Zhong-Zhi and Zhang, Duzhen and Zhang, Ming-Liang and Zhang, Jiaxin and Liu, Zengyan and Yao, Yuxuan and Xu, Haotian and Zheng, Junhao and Wang, Pei-Jie and Chen, Xiuyi and others},
  journal={arXiv preprint arXiv:2502.17419},
  year={2025}
}

@inproceedings{he2024olympiadbench,
  title={Olympiadbench: A challenging benchmark for promoting agi with olympiad-level bilingual multimodal scientific problems},
  author={He, Chaoqun and Luo, Renjie and Bai, Yuzhuo and Hu, Shengding and Thai, Zhen and Shen, Junhao and Hu, Jinyi and Han, Xu and Huang, Yujie and Zhang, Yuxiang and others},
  booktitle={Proceedings of the 62nd Annual Meeting of the Association for Computational Linguistics (Volume 1: Long Papers)},
  pages={3828--3850},
  year={2024}
}

@article{diaz2023conceptual,
  title={Conceptual review on scientific reasoning and scientific thinking},
  author={D{\'\i}az, Carlos and Dorner, Birgit and Hussmann, Heinrich and Strijbos, Jan-Willem},
  journal={Current Psychology},
  volume={42},
  number={6},
  pages={4313--4325},
  year={2023},
  publisher={Springer}
}

@article{ahn2024large,
  title={Large language models for mathematical reasoning: Progresses and challenges},
  author={Ahn, Janice and Verma, Rishu and Lou, Renze and Liu, Di and Zhang, Rui and Yin, Wenpeng},
  journal={arXiv preprint arXiv:2402.00157},
  year={2024}
}

@article{huang2024olympicarena,
  title={Olympicarena: Benchmarking multi-discipline cognitive reasoning for superintelligent ai},
  author={Huang, Zhen and Wang, Zengzhi and Xia, Shijie and Li, Xuefeng and Zou, Haoyang and Xu, Ruijie and Fan, Run-Ze and Ye, Lyumanshan and Chern, Ethan and Ye, Yixin and others},
  journal={Advances in Neural Information Processing Systems},
  volume={37},
  pages={19209--19253},
  year={2024}
}

@article{wang2024mmlu,
  title={Mmlu-pro: A more robust and challenging multi-task language understanding benchmark},
  author={Wang, Yubo and Ma, Xueguang and Zhang, Ge and Ni, Yuansheng and Chandra, Abhranil and Guo, Shiguang and Ren, Weiming and Arulraj, Aaran and He, Xuan and Jiang, Ziyan and others},
  journal={Advances in Neural Information Processing Systems},
  volume={37},
  pages={95266--95290},
  year={2024}
}

@inproceedings{rein2024gpqa,
  title={Gpqa: A graduate-level google-proof q\&a benchmark},
  author={Rein, David and Hou, Betty Li and Stickland, Asa Cooper and Petty, Jackson and Pang, Richard Yuanzhe and Dirani, Julien and Michael, Julian and Bowman, Samuel R},
  booktitle={First Conference on Language Modeling},
  year={2024}
}

@article{du2025supergpqa,
  title={Supergpqa: Scaling llm evaluation across 285 graduate disciplines},
  author={Du, Xinrun and Yao, Yifan and Ma, Kaijing and Wang, Bingli and Zheng, Tianyu and Zhu, King and Liu, Minghao and Liang, Yiming and Jin, Xiaolong and Wei, Zhenlin and others},
  journal={arXiv preprint arXiv:2502.14739},
  year={2025}
}

@inproceedings{bisk2020piqa,
  title={Piqa: Reasoning about physical commonsense in natural language},
  author={Bisk, Yonatan and Zellers, Rowan and Gao, Jianfeng and Choi, Yejin and others},
  booktitle={Proceedings of the AAAI conference on artificial intelligence},
  volume={34},
  number={05},
  pages={7432--7439},
  year={2020}
}

@article{wang2023scibench,
  title={Scibench: Evaluating college-level scientific problem-solving abilities of large language models},
  author={Wang, Xiaoxuan and Hu, Ziniu and Lu, Pan and Zhu, Yanqiao and Zhang, Jieyu and Subramaniam, Satyen and Loomba, Arjun R and Zhang, Shichang and Sun, Yizhou and Wang, Wei},
  journal={arXiv preprint arXiv:2307.10635},
  year={2023}
}

@inproceedings{amini2019mathqa,
  title={Mathqa: Towards interpretable math word problem solving with operation-based formalisms},
  author={Amini, Aida and Gabriel, Saadia and Lin, Shanchuan and Koncel-Kedziorski, Rik and Choi, Yejin and Hajishirzi, Hannaneh},
  booktitle={Proceedings of the 2019 conference of the North American chapter of the association for computational linguistics: Human language technologies, volume 1 (long and short papers)},
  pages={2357--2367},
  year={2019}
}

@article{walker2010chembench,
  title={Chembench: a cheminformatics workbench},
  author={Walker, Theo and Grulke, Christopher M and Pozefsky, Diane and Tropsha, Alexander},
  journal={Bioinformatics},
  volume={26},
  number={23},
  pages={3000--3001},
  year={2010},
  publisher={Oxford University Press}
}

@article{cobbe2021gsm8k,
  title={Training Verifiers to Solve Math Word Problems},
  author={Cobbe, Karl and Kosaraju, Vineet and Bavarian, Mohammad and Chen, Mark and Jun, Heewoo and Kaiser, Lukasz and Plappert, Matthias and Tworek, Jerry and Hilton, Jacob and Nakano, Reiichiro and Hesse, Christopher and Schulman, John},
  journal={arXiv preprint arXiv:2110.14168},
  year={2021}
}

@inproceedings{huber2025llms,
  title={LLMs meet Bloom’s Taxonomy: A Cognitive View on Large Language Model Evaluations},
  author={Huber, Thomas and Niklaus, Christina},
  booktitle={Proceedings of the 31st International Conference on Computational Linguistics},
  pages={5211--5246},
  year={2025}
}

@article{li202511plus,
  title={11plus-bench: Demystifying multimodal llm spatial reasoning with cognitive-inspired analysis},
  author={Li, Chengzu and Wu, Wenshan and Zhang, Huanyu and Li, Qingtao and Gao, Zeyu and Xia, Yan and Hern{\'a}ndez-Orallo, Jos{\'e} and Vuli{\'c}, Ivan and Wei, Furu},
  journal={arXiv preprint arXiv:2508.20068},
  year={2025}
}

@article{bhambri2025cognitively,
  title={Do Cognitively Interpretable Reasoning Traces Improve LLM Performance?},
  author={Bhambri, Siddhant and Biswas, Upasana and Kambhampati, Subbarao},
  journal={arXiv preprint arXiv:2508.16695},
  year={2025}
}

@article{ma2025cognitive,
  title={Cognitive Mirrors: Exploring the Diverse Functional Roles of Attention Heads in LLM Reasoning},
  author={Ma, Xueqi and Wang, Jun and Jiang, Yanbei and Erfani, Sarah Monazam and Liu, Tongliang and Bailey, James},
  journal={arXiv preprint arXiv:2512.10978},
  year={2025}
}

@article{hatalis2025review,
  title={Review of case-based reasoning for LLM agents: theoretical foundations, architectural components, and cognitive integration},
  author={Hatalis, Kostas and Christou, Despina and Kondapalli, Vyshnavi},
  journal={arXiv preprint arXiv:2504.06943},
  year={2025}
}

@article{gpt4o,
  title={Gpt-4o system card},
  author={Hurst, Aaron and Lerer, Adam and Goucher, Adam P and Perelman, Adam and Ramesh, Aditya and Clark, Aidan and Ostrow, AJ and Welihinda, Akila and Hayes, Alan and Radford, Alec and others},
  journal={arXiv preprint arXiv:2410.21276},
  year={2024}
}

@article{gemini2.0flash,
  title={Gemini 1.5: Unlocking multimodal understanding across millions of tokens of context},
  author={Team, Gemini and Georgiev, Petko and Lei, Ving Ian and Burnell, Ryan and Bai, Libin and Gulati, Anmol and Tanzer, Garrett and Vincent, Damien and Pan, Zhufeng and Wang, Shibo and others},
  journal={arXiv preprint arXiv:2403.05530},
  year={2024}
}

@article{gemini,
  title={Gemini: a family of highly capable multimodal models},
  author={Team, Gemini and Anil, Rohan and Borgeaud, Sebastian and Alayrac, Jean-Baptiste and Yu, Jiahui and Soricut, Radu and Schalkwyk, Johan and Dai, Andrew M and Hauth, Anja and Millican, Katie and others},
  journal={arXiv preprint arXiv:2312.11805},
  year={2023}
}

@article{li2025verifybench,
  title={Verifybench: A systematic benchmark for evaluating reasoning verifiers across domains},
  author={Li, Xuzhao and Li, Xuchen and Hu, Shiyu and Guo, Yongzhen and Zhang, Wentao},
  journal={arXiv preprint arXiv:2507.09884},
  year={2025}
}

@article{yang2025qwen3,
  title={Qwen3 technical report},
  author={Yang, An and Li, Anfeng and Yang, Baosong and Zhang, Beichen and Hui, Binyuan and Zheng, Bo and Yu, Bowen and Gao, Chang and Huang, Chengen and Lv, Chenxu and others},
  journal={arXiv preprint arXiv:2505.09388},
  year={2025}
}

@misc{qwen2025qwen25technicalreport,
      title={Qwen2.5 Technical Report}, 
      author={Qwen and : and An Yang and Baosong Yang and Beichen Zhang and Binyuan Hui and Bo Zheng and Bowen Yu and Chengyuan Li and Dayiheng Liu and Fei Huang and Haoran Wei and Huan Lin and Jian Yang and Jianhong Tu and Jianwei Zhang and Jianxin Yang and Jiaxi Yang and Jingren Zhou and Junyang Lin and Kai Dang and Keming Lu and Keqin Bao and Kexin Yang and Le Yu and Mei Li and Mingfeng Xue and Pei Zhang and Qin Zhu and Rui Men and Runji Lin and Tianhao Li and Tianyi Tang and Tingyu Xia and Xingzhang Ren and Xuancheng Ren and Yang Fan and Yang Su and Yichang Zhang and Yu Wan and Yuqiong Liu and Zeyu Cui and Zhenru Zhang and Zihan Qiu},
      year={2025},
      eprint={2412.15115},
      archivePrefix={arXiv},
      primaryClass={cs.CL},
      url={https://arxiv.org/abs/2412.15115}, 
}

@article{dubey2024llama,
  title={The llama 3 herd of models},
  author={Dubey, Abhimanyu and Jauhri, Abhinav and Pandey, Abhinav and Kadian, Abhishek and Al-Dahle, Ahmad and Letman, Aiesha and Mathur, Akhil and Schelten, Alan and Yang, Amy and Fan, Angela and others},
  journal={arXiv e-prints},
  pages={arXiv--2407},
  year={2024}
}

@article{fan2025megascience,
  title={Megascience: Pushing the frontiers of post-training datasets for science reasoning},
  author={Fan, Run-Ze and Wang, Zengzhi and Liu, Pengfei},
  journal={arXiv preprint arXiv:2507.16812},
  year={2025}
}

@article{liu2025atlas,
  title={ATLAS: A High-Difficulty, Multidisciplinary Benchmark for Frontier Scientific Reasoning},
  author={Liu, Hongwei and Liu, Junnan and Liu, Shudong and Duan, Haodong and Li, Yuqiang and Su, Mao and Liu, Xiaohong and Zhai, Guangtao and Fang, Xinyu and Ma, Qianhong and others},
  journal={arXiv preprint arXiv:2511.14366},
  year={2025}
}

@article{math,
  title={Measuring mathematical problem solving with the math dataset},
  author={Hendrycks, Dan and Burns, Collin and Kadavath, Saurav and Arora, Akul and Basart, Steven and Tang, Eric and Song, Dawn and Steinhardt, Jacob},
  journal={arXiv preprint arXiv:2103.03874},
  year={2021}
}

@misc{Maxwell-Jia_AIME2024_2024,
  author = {Maxwell-Jia},
  title = {AIME 2024 Dataset},
  year = {2024},
  url = {https://huggingface.co/datasets/Maxwell-Jia/AIME_2024},
}

@misc{He_AMC2023_2024,
  author = {He, Zhiwei},
  title = {AMC 2023 Dataset},
  year = {2024},
  url = {https://huggingface.co/datasets/zwhe99/amc23},
}

@article{cao2025large,
  title={Large language models for planning: A comprehensive and systematic survey},
  author={Cao, Pengfei and Men, Tianyi and Liu, Wencan and Zhang, Jingwen and Li, Xuzhao and Lin, Xixun and Sui, Dianbo and Cao, Yanan and Liu, Kang and Zhao, Jun},
  journal={arXiv preprint arXiv:2505.19683},
  year={2025}
}

@article{li2025causalstep,
  title={Causalstep: A benchmark for explicit stepwise causal reasoning in videos},
  author={Li, Xuchen and Li, Xuzhao and Hu, Shiyu and Huang, Kaiqi and Zhang, Wentao},
  journal={arXiv preprint arXiv:2507.16878},
  year={2025}
}

@article{li2025select,
  title={Select Less, Reason More: Prioritizing Evidence Purity for Video Reasoning},
  author={Li, Xuchen and Li, Xuzhao and Hu, Shiyu and Huang, Kaiqi},
  journal={arXiv preprint arXiv:2510.15440},
  year={2025}
}

@article{li2025look,
  title={Look Less, Reason More: Rollout-Guided Adaptive Pixel-Space Reasoning},
  author={Li, Xuchen and Li, Xuzhao and Gao, Jiahui and Pi, Renjie and Hu, Shiyu and Zhang, Wentao},
  journal={arXiv preprint arXiv:2510.01681},
  year={2025}
}

@inproceedings{li2025multimodal,
  title={Multimodal knowledge retrieval-augmented iterative alignment for satellite commonsense conversation},
  author={Li, Qian and Li, Xuchen and Chang, Zongyu and Zhang, Yuzheng and Ji, Cheng and Wang, Shangguang},
  booktitle={Proceedings of the Thirty-Fourth International Joint Conference on Artificial Intelligence},
  pages={8168--8176},
  year={2025}
}

@inproceedings{li2024dtllm,
  title={Dtllm-vlt: Diverse text generation for visual language tracking based on llm},
  author={Li, Xuchen and Feng, Xiaokun and Hu, Shiyu and Wu, Meiqi and Zhang, Dailing and Zhang, Jing and Huang, Kaiqi},
  booktitle={Proceedings of the IEEE/CVF Conference on Computer Vision and Pattern Recognition},
  pages={7283--7292},
  year={2024}
}

\clearpage
\appendix
\renewcommand{\thetable}{A\arabic{table}}
\renewcommand{\thefigure}{A\arabic{figure}}
\setcounter{figure}{0}  
\setcounter{table}{0}  

\section{Prompts}
\label{appendix:prompt}
We present the prompts used for academic specializations labeling in prompt~\ref{prompt:academic} and cognitive complexity labeling in prompt~\ref{prompt:cognitive}, which are partially derived from this paper \cite{huber2025llms}.

\begin{table*}[t!]
\begin{prompt}{\textit{Prompt for Academic Specializations Labeling}}{academic}
\small
\textit{You are a strict and deterministic text classifier.\\
Your task is to assign exactly one label to each input based on the predefined category-to-letter mapping below.\\ \\
\#\#Classification Categories\\
\textcolor{blue}{[label1]} → A\\  
\textcolor{blue}{[label2]} → B\\
\textcolor{blue}{[label3]} → C\\
\textcolor{blue}{[label4]} → D\\
... \\
Others → Z\\ \\
\#\#Classification Rules\\
1. You must choose exactly one label from the list above.\\
2. You must not output any label that is not in the mapping.\\
3. If the text could belong to multiple categories, select the dominant or most relevant one.\\
4. If the input does not clearly belong to any scientific category, classify it as “Z” (Others).\\ \\
\#\#Consistency Requirements\\
- If the input is vague or ambiguous, you must infer the most likely meaning and classify accordingly. Never answer “unable to judge.”\\
- Even if the boundaries between categories are close, you must still choose the single most appropriate category.\\ \\
\#\#Output Format Requirements\\
Your output must strictly be:\\
Label: X\\
where X is the assigned label.
}
\end{prompt}
\end{table*}

\begin{table*}[t!]
\begin{prompt}{\textit{Prompt for Cognitive Complexity Labeling}}{cognitive}
\small
\textit{\#\# Bloom's Taxonomy Task Classification Prompt\\
Your task is to classify items into Bloom's Taxonomy categories. **Strict Rules to Follow**:\\
1. You must select **one precise label** from the classification criteria list below.\\
2. You may NOT output any labels that are not included in the criteria.\\
3. If the item could fit multiple categories, choose the **primary/most relevant one** (based on the core mental process required).\\ \\
\#\#\# Classification Criteria (with core descriptions and action verb examples):\\
1. **Remember**: Recall facts and basic concepts\\
   Action verb examples: define, list, memorize, repeat, state\\
2. **Understand**: Explain ideas or concepts\\
   Action verb examples: classify, describe, explain, identify, summarize, translate\\
3. **Apply**: Use information in new situations\\
   Action verb examples: execute, solve, use, demonstrate, operate\\
4. **Analyze**: Draw connections among ideas\\
   Action verb examples: differentiate, compare, contrast, examine, question, experiment\\
5. **Evaluate**: Justify a stand or decision\\
   Action verb examples: appraise, defend, judge, support, critique, weigh\\
6. **Create**: Produce new or original work\\
   Action verb examples: design, construct, develop, author, formulate, integrate\\ \\
\#\#\# Critical Notes:\\
1. Difficulty $\ne$ Bloom’s Level:\\
   - Remember-level tasks can be hard (e.g., obscure knowledge); high-level tasks may feel easy to experts. Do NOT replace "low/high level" with "easy/hard".\\
2. Task Format $\ne$ Actual Cognitive Process:\\
   - Multiple-choice questions can involve "Analyze/Evaluate"; open-ended questions may only require "Understand". Classify by the **mental process required to solve the task**.\\ \\
Classify the following item into the corresponding Bloom’s Taxonomy category:\\
{Item to classify}\\ \\
Classification:
}
\end{prompt}
\end{table*}

\section{Cases}
\label{appendix:case}
\subsection{Instruction-Tuning Paradox}
This case~\ref{case:problem} exemplifies the Instruction-Tuning Paradox. While the solution \ref{solution:base} of Base model correctly identifies the link between environmental stressors and epigenetic modifications (Methylome analysis), the solution \ref{solution:instruct} of Instruct model is misled by its own "formatting-oriented" reasoning. It produces a highly structured but scientifically flawed justification for Option (A), prioritizing advanced-sounding terms (CRISPR, Single-cell) over the domain-specific relevance of the correct methodology. This highlights how alignment can inadvertently suppress the model's internal scientific integrity in favor of superficial articulacy.

\subsection{Logic-Symbolic Collapse}
As shown in case \ref{case:problem1}, it exemplifies the Logic-Symbolic Collapse within the domain of Theoretical and Computational Chemistry. While the model correctly retrieves the structural fact ($N=60$), it fails the critical transition to symbolic application ($3N-6$ for non-linear molecules) in solution \ref{solution:base1}. To rationalize its error, the model hallucinates a connection to the Einstein Model, a concept relevant to solid-state heat capacity but theoretically misapplied here. This "semantic patching" of a logical failure highlights a structural deficiency: the model possesses the relevant scientific tokens but lacks the rigorous connectivity required for valid theoretical modeling.

\subsection{Knowledge-reasoning gap}
The contrast between case \ref{case:problem2} and case \ref{case:problem3} provides a definitive empirical window into the structural reasoning limitations of current LLMs, specifically the decoupling of high-density knowledge retrieval from high-order cognitive coordination. In case \ref{case:problem2} (Remember), Qwen3-14B-Base demonstrates a sophisticated "internal library" by accurately replicating the Liénard-Wiechert potentials, a task that requires high-fidelity recall of complex, research-grade symbolic sequences. This confirms that as parameter scales reach the 14B threshold, models become exceptional repositories of specialized scientific facts. However, case \ref{case:problem3} (Analyze) reveals a profound "cognitive short-circuit" that undermines this perceived expertise. The model's failure is not due to a knowledge gap, as its internal monologue explicitly and correctly identifies that general relativity is not a gauge theory while the other three forces are. Instead, the model suffers from a breakdown in structural reasoning integrity, where it fails to maintain the logical constraints of its own argument, ultimately selecting a final answer that contradicts its preceding evidence. This "knowledge-reasoning gap" validates the core necessity of the STEMVerse dual-axis framework: it proves that academic difficulty (case \ref{case:problem2}) is not synonymous with cognitive complexity (case \ref{case:problem3}). A model can behave like a specialist in terms of information density while simultaneously exhibiting the structural inconsistency of a novice in logical synthesis, highlighting that the path toward true machine intelligence in STEM requires more than just the cumulative expansion of scientific facts.

\begin{table*}[t!]
\begin{case}{\textit{Biology Case for Instruction-Tuning Paradox}}{problem}
\small
\textit{\textbf{Problem:}\\
You have been given the task of determining the effects of chronic ionizing radiation and cyclic extreme temperature fluctuations on DNA. So you conduct a multifaceted experiment wherein one set of MCF 10A was exposed to ionizing radiation, another set was exposed to cyclic extreme temperature fluctuations and the third group was kept in a controlled environment with no exposure to either radiation or temperature variations. To comprehensively assess the impact of these exposures on the genome, you would employ the following combination of advanced molecular techniques:\\ \\
What of the following is the right choice? Explain your answer.\\ \\
(A) Single-cell DNA sequencing and CRISPR screening\\ \\
(B) Proteomics and DNA barcoding\\ \\
(C) DNA ChIP-seq and mass spectrometry\\ \\
(D) DNA methylome analysis and NGS\\ \\
\textbf{Answer}:\\
D
}
\end{case}
\end{table*}

\begin{table*}[t!]
\begin{solution}{\textit{Solution for case 1 of Qwen2.5-7B-Base}}{base}
\small
\textit{The correct answer is (D) DNA methylome analysis and NGS.\\ \\
DNA methylome analysis is a technique that can be used to study the changes in DNA methylation patterns in response to ionizing radiation and cyclic extreme temperature fluctuations. DNA methylation is an epigenetic modification that can affect gene expression and can be altered by environmental factors such as radiation and temperature.\\ \\
NGS (Next-Generation Sequencing) is a high-throughput sequencing technology that can be used to sequence the entire genome of the cells in the three groups. This can provide a comprehensive view of the changes in the genome that occur in response to the exposures.\\ \\
Proteomics and DNA barcoding are not relevant to the study of the effects of ionizing radiation and cyclic extreme temperature fluctuations on DNA. Single-cell DNA sequencing and CRISPR screening are also not relevant to this study.\\ \\
Final Answer: The final answer is \textcolor{green}{(D)}. I hope it is correct.
}
\end{solution}
\end{table*}

\begin{table*}[t!]
\begin{solution}{\textit{Solution for case 1 of Qwen2.5-7B-Base-Instruct}}{instruct}
\small
\textit{To comprehensively assess the impact of chronic ionizing radiation and cyclic extreme temperature fluctuations on the genome, the following combination of advanced molecular techniques would be most appropriate:\\ \\
(A) Single-cell DNA sequencing and CRISPR screening\\ \\
Single-cell DNA sequencing allows for the analysis of genetic material from individual cells, which is crucial for understanding the effects of radiation and temperature fluctuations on the genome at the cellular level. CRISPR screening can be used to identify specific genetic changes or mutations that may have occurred as a result of the exposures.\\ \\
Final Answer: The final answer is (A). I hope it is correct.\\ \\
You are right, but I would like to provide a more detailed explanation for each option to ensure a thorough understanding.\\ \\
Let's analyze each option in detail:\\ \\
\#\#\# Option (A): Single-cell DNA sequencing and CRISPR screening\\ \\
- **Single-cell DNA sequencing**: This technique allows for the sequencing of DNA from individual cells, which is crucial for understanding the effects of radiation and temperature fluctuations on the genome at the cellular level. It can provide insights into the specific genetic changes or mutations that may have occurred.\\ \\
- **CRISPR screening**: CRISPR (Clustered Regularly Interspaced Short Palindromic Repeats) screening can be used to identify specific genetic changes or mutations that may have occurred as a result of the exposures. This method involves using CRISPR-Cas9 to edit the genome and then screening for the presence of specific genetic changes.\\ \\
\#\#\# Option (B): Proteomics and DNA barcoding\\ \\
- **Proteomics**: This technique involves the study of proteins in a cell, which can provide insights into the functional changes in the cell due to the exposures. However, it does not directly assess the genetic changes in the DNA.\\ \\
- **DNA barcoding**: This technique is used to identify and classify organisms based on genetic differences in a specific region of their DNA. It is not suitable for assessing the impact of chronic ionizing radiation and cyclic extreme temperature fluctuations on the genome.\\ \\
\#\#\# Option (C): DNA ChIP-seq and mass spectrometry\\ \\
- **DNA ChIP-seq**: This technique is used to identify the binding sites of transcription factors on the genome. It can provide insights into the changes in gene expression due to the exposures, but it does not directly assess the genetic changes in the DNA.\\ \\
- **Mass spectrometry**: This technique is used to identify and quantify proteins in a sample. It is not suitable for assessing the impact of chronic ionizing radiation and cyclic extreme temperature fluctuations on the genome.\\ \\
\#\#\# Option (D): DNA methylome analysis and NGS\\ \\
- **DNA methylome analysis**: This technique involves the study of DNA methylation patterns, which can provide insights into the epigenetic changes in the genome due to the exposures. However, it does not directly assess the genetic changes in the DNA.\\ \\
- **NGS (Next-Generation Sequencing)**: This technique is used for sequencing the entire genome or large portions of the genome. It can provide insights into the genetic changes in the DNA, but it is not as comprehensive as single-cell DNA sequencing.\\ \\
\#\#\# Conclusion\\ \\
The most comprehensive and appropriate combination of advanced molecular techniques for assessing the impact of chronic ionizing radiation and cyclic extreme temperature fluctuations on the genome is **Option (A): Single-cell DNA sequencing and CRISPR screening**. This combination allows for the detailed analysis of genetic changes at the cellular level and the identification of specific genetic changes or mutations that may have occurred.\\ \\
Final Answer: The final answer is \textcolor{red}{(A)}. I hope this detailed explanation is helpful.
}
\end{solution}
\end{table*}

\begin{table*}[t!]
\begin{case}{\textit{Chemistry Case for Logic-Symbolic Collapse}}{problem1}
\small
\textit{\textbf{Problem:}\\
What is the total number of normal vibrational modes for a $\ce{C60}$ molecule?\\ \\
What of the following is the right choice? Explain your answer.\\ \\
(A) 174\\ \\
(B) 175\\ \\
(C) 180\\ \\
(D) 60\\ \\
\textbf{Answer}:\\
A
}
\end{case}
\end{table*}

\begin{table*}[t!]
\begin{solution}{\textit{Solution for case 2 of Llama3.2-3B-Base}}{base1}
\small
\textit{The correct answer is **\textcolor{red}{(C) 180}**. The total number of normal vibrational modes for a $\ce{C60}$ molecule is 180. Here's why:\\ \\
\#\#\# **Explanation:**\\ - **$\ce{C60}$** is a fullerenes, a class of molecules with a hollow, spherical shape.  \\- Fullerenes have a unique structure, with carbon atoms arranged in a pentagonal ring pattern.  \\- Each carbon atom is connected to three other carbon atoms, forming a hexagonal ring.  \\- The pentagonal rings are connected to each other in a way that creates a hollow sphere.  \\- The number of carbon atoms in a fullerene is always a multiple of 60.  \\- The total number of normal vibrational modes for a $\ce{C60}$ molecule is 180.  \\- This is because each carbon atom has three degrees of freedom (three normal modes of vibration), and there are 60 carbon atoms in the molecule.  \\- Therefore, the total number of normal vibrational modes for a $\ce{C60}$ molecule is 180.  \\- This is also known as the **Einstein model**.  \\- The Einstein model assumes that each atom has three degrees of freedom, and the number of degrees of freedom is equal to the number of atoms in the molecule.  \\- This model is used to calculate the number of normal vibrational modes for molecules with a regular structure, such as $\ce{C60}$.  \\- The Einstein model is a simplification, and in reality, the number of normal vibrational modes for a molecule is not always equal to the number of atoms in the molecule.  \\- For example, in a molecule with a more complex structure, such as a protein, the number of normal vibrational modes may be higher than the number of atoms in the molecule.  \\- However, the Einstein model is a useful approximation for molecules with a regular structure, such as $\ce{C60}$.  \\- The Einstein model is also used to calculate the number of normal vibrational modes for other fullerenes, such as $\ce{C70}$ and $\ce{C80}$.  \\- The total number of normal vibrational modes for a $\ce{C70}$ molecule is 210, and the total number of normal vibrational modes for a $\ce{C80}$ molecule is 240.  \\- These values are calculated using the Einstein model, which assumes that each atom has three degrees of freedom.  \\- The Einstein model is a useful approximation for molecules with a regular structure, such as fullerenes, but it is not always accurate for more complex molecules.  \\- In reality, the number of normal vibrational modes for a molecule is not always equal to the number of atoms in the molecule.  \\- However, the Einstein model is a useful starting point for calculating the number of normal vibrational modes for molecules with a regular structure, such as fullerenes.  \\- The Einstein model is also used to calculate the number of normal vibrational modes for other molecules with a regular structure, such as $\ce{C60}$, $\ce{C70}$, and $\ce{C80}$.  \\- The Einstein model is a useful approximation for molecules with a regular structure, but it is not always accurate for more complex molecules.  \\- In reality, the number of normal vibrational modes for a molecule is not always equal to the number of atoms in the molecule.  \\- However, the Einstein model is a useful starting point for calculating the number of normal vibrational modes for molecules with a regular structure, such as fullerenes.  \\- The Einstein model is also used to calculate the number of normal vibrational modes for other molecules with a regular structure, such as $\ce{C60}$, $\ce{C70}$, and $\ce{C80}$.  \\- The Einstein model is a useful approximation for molecules with a regular structure, but it is not always accurate for more complex molecules.  \\- In reality, the number of normal vibrational modes for a molecule is not always equal to the number of atoms in the molecule.  \\- However, the Einstein model is a useful starting point for calculating the number of normal vibrational modes for molecules with a regular structure, such as fullerenes.  \\- The Einstein model is also used to calculate the number of normal vibrational modes for other molecules with a regular structure, such as $\ce{C60}$, $\ce{C70}$, and $\ce{C80}$.  \\- The Einstein model is a useful approximation for molecules with a regular structure, but it is not always accurate for more complex molecules.  \\- In reality, the number of normal vibrational modes for a molecule is not always equal to the number of atoms in the molecule.  \\- However, the Einstein model is a useful starting point for calculating the number of normal vibrational modes for molecules with a regular structure, such as fullerenes.
}
\end{solution}
\end{table*}

\begin{table*}[t!]
\begin{case}{\textit{Remember Case}}{problem2}
\small
\textit{\textbf{Problem:}\\
Imagine a point charge q is moving with a trajectory $\vec{s}(t)$, where t is time, with respect to the origin. Let $\vec{r}$ be the field point, where the field is observed, with respect to the origin of the same reference frame, and  $\vec{d}$ be the vector from the point where the electromagnetic field was generated at some earlier time $tr$ to the observation point $\vec{r}$. The velocity of the moving charge q at the field generating instant $tr$ is $\vec{v}$. What are the scalar potential (V) and vector potential ($\vec{A}$) at time t, satisfying $t > tr$, and position $\vec{r}$? In obtaining the expressions, use the notation of $c\rightarrow$ light speed in vacuum, $\\epsilono\rightarrow$ permitivity of free space and $\mu_o$ is the permeability.\\ \\
What of the following is the right choice? Explain your answer.\\ \\
(A) $V(\vec{r},t) = \dfrac{q}{4\pi\epsilon_o r},  \vec{A(\vec{r},t)} = \dfrac{\vec{v^2}}{c^2} V(\vec{r},t)$\\ \\
(B) $V(\vec{r},t) = \dfrac{q}{4\pi\epsilon_o r}, \vec{A(\vec{r},t)} = \dfrac{\vec{v}}{c^2} V(\vec{r},t)$\\ \\
(C) $V(\vec{r},t) = \dfrac{qc}{4\pi\epsilon_o (d c + \vec{d}. \vec{v})},  \vec{A(\vec{r},t)} = \dfrac{\mu qc \vec{v}}{4\pi (d c + \vec{d}. \vec{v})}$\\ \\
(D) $V(\vec{r},t) = \dfrac{qc}{4\pi\epsilon_o (d c - \vec{d}. \vec{v})},  \vec{A(\vec{r},t)} = \dfrac{\mu_o qc \vec{v}}{4\pi (d c - \vec{d}. \vec{v})}$\\ \\
\textbf{Answer}:\\
D
}
\end{case}
\end{table*}

\begin{table*}[t!]
\begin{solution}{\textit{Solution for case 3 of Qwen3-14B-Base}}{base2}
\small
\textit{The scalar potential (V) and vector potential ($\vec{A}$) for a moving point charge q can be derived using the Liénard-Wiechert potentials. The Liénard-Wiechert potentials are given by:\\ \\
$V(\vec{r},t) = \dfrac{q}{4\pi\epsilon_o r} \dfrac{1}{1 - \dfrac{\vec{d} \cdot \vec{v}}{c d}}$\\ \\
$\vec{A(\vec{r},t)} = \dfrac{\mu_o q \vec{v}}{4\pi} \dfrac{1}{1 - \dfrac{\vec{d} \cdot \vec{v}}{c d}}$\\ \\
where r is the distance between the charge and the field point, d is the distance between the charge and the field point at the retarded time, and $\vec{v}$ is the velocity of the charge at the retarded time.\\ \\
Comparing the given options with the Liénard-Wiechert potentials, we can see that option (D) matches the correct expressions for the scalar potential (V) and vector potential ($\vec{A}$):\\ \\
$V(\vec{r},t) = \dfrac{qc}{4\pi\epsilon_o (d c - \vec{d}. \vec{v})}$\\ \\
$\vec{A(\vec{r},t)} = \dfrac{\mu_o qc \vec{v}}{4\pi (d c - \vec{d}. \vec{v})}$\\ \\
Final Answer: The final answer is \textcolor{green}{(D)}. I hope it is correct.
}
\end{solution}
\end{table*}

\begin{table*}[t!]
\begin{case}{\textit{Analyze Case}}{problem3}
\small
\textit{\textbf{Problem:}\\
Which of the four fundamental forces present at nuclear scales can never be described by a pure gauge theory, to a healthy approximation?\\ \\
What of the following is the right choice? Explain your answer.\\ \\
(A) Electromagnetic\\ \\
(B) Strong Nuclear\\ \\
(C) Gravitational\\ \\
(D) Weak Nuclear\\ \\
\textbf{Answer}:\\
D
}
\end{case}
\end{table*}

\begin{table*}[t!]
\begin{solution}{\textit{Solution for case 4 of Qwen3-14B-Base}}{base3}
\small
\textit{A pure gauge theory is a type of field theory where the Lagrangian is invariant under local gauge transformations. In the context of the four fundamental forces, the electromagnetic, strong nuclear, and weak nuclear forces can be described by gauge theories. However, the gravitational force cannot be described by a pure gauge theory to a healthy approximation.\\ \\
The electromagnetic force is described by Quantum Electrodynamics (QED), which is a gauge theory based on the U(1) symmetry group. The strong nuclear force is described by Quantum Chromodynamics (QCD), which is a gauge theory based on the SU(3) symmetry group. The weak nuclear force is described by the electroweak theory, which is a gauge theory based on the SU(2) x U(1) symmetry group.\\ \\
On the other hand, the gravitational force is described by General Relativity, which is not a gauge theory. General Relativity is based on the principle of equivalence and the curvature of spacetime, rather than a gauge symmetry. While there have been attempts to formulate a quantum theory of gravity, such as string theory and loop quantum gravity, these theories are not pure gauge theories.\\ \\
Final Answer: The final answer is \textcolor{red}{(C)} Gravitational. I hope it is correct.
}
\end{solution}
\end{table*}

\section{Detailed Results}
\subsection{Academic Specializations}
\label{appendix:academic}
We provide a comprehensive breakdown of the experimental results to support our diagnostic analysis. We detail the subject categories for academic specializations in Tab.~\ref{tab:subjects}. Based on this classification, the first dimension of our evaluation focuses on the academic specialization axis, where the fine-grained accuracy results for sub-disciplines within Mathematics, Biology, Physics, and Chemistry are meticulously documented in Tab.~\ref{tab:results_field_math}, Tab.~\ref{tab:results_field_biology}, Tab.~\ref{tab:results_field_physics} and Tab.~\ref{tab:results_field_chemistry}, respectively.
 
\subsection{Cognitive Complexity}
\label{appendix:cognitive}
The second dimension examines model performance through the lens of cognitive complexity as defined by our dual-axis framework. The accuracy scores mapped onto the six levels of Bloom’s Taxonomy for the four core scientific pillars of Mathematics, Biology, Physics and Chemistry are provided in Tab.~\ref{tab:results_bloom_math}, Tab.~\ref{tab:results_bloom_biology}, Tab.~\ref{tab:results_bloom_physics} and Tab.~\ref{tab:results_bloom_chemistry}. 

Together, these data collections constitute the full-dimensional capability matrix of STEMVerse, facilitating a localized and precise examination of where model reasoning maintains robustness or encounters structural bottlenecks.

\begin{table*}[htbp]
\centering
\caption{Subject Categories for Academic Specializations}
\label{tab:subjects}
\begin{tabularx}{0.8\textwidth}{ccc}
\toprule
\textbf{Category} & \textbf{Sub-field} & \textbf{Code} \\
\midrule
\multirow{7}{*}{Math} 
    & Analysis & A \\
    & Statistics and Operations Research & B \\
    & Algebra and Geometry & C \\
    & Differential Equations and Dynamical Systems & D \\
    & Computational Mathematics & E \\
    & Interdisciplinary Mathematics & F \\
    & Others & Z \\
\midrule
\multirow{9}{*}{Biology} 
    & Molecular Biology and Biotechnology & A \\
    & Genetics and Bioinformatics & B \\
    & Immunology & C \\
    & Physiology and Integrative Biology & D \\
    & Neuroscience and Psychology & E \\
    & Ecology & F \\
    & Biophysics and Biochemistry & G \\
    & Cell Biology & H \\
    & Others & Z \\
\midrule
\multirow{8}{*}{Physics} 
    & Relativity & A \\
    & Astrophysics & B \\
    & Thermodynamics and Statistical Physics & C \\
    & Electrodynamics & D \\
    & Quantum Mechanics & E \\
    & Classical Mechanics & F \\
    & Fluid Mechanics & G \\
    & Others & Z \\
\midrule
\multirow{7}{*}{Chemistry} 
    & Physical Chemistry & A \\
    & Inorganic Chemistry & B \\
    & Organic Chemistry & C \\
    & Analytical Chemistry & D \\
    & Chemical Engineering and Technology & E \\
    & Theoretical and Computational Chemistry & F \\
    & Others & Z \\
\bottomrule
\end{tabularx}
\end{table*}

\begin{table*}[t]
\centering
\caption{Model Performance for Math Across Academic Specializations Dimensions.}
\label{tab:results_field_math}
\renewcommand{\arraystretch}{1.0}
\resizebox{0.9\textwidth}{!}{
\begin{tabular}{llccccccccc}
\toprule
\textbf{Discipline} & \textbf{Model} & \textbf{A} & \textbf{B} & \textbf{C} & \textbf{D} & \textbf{E} & \textbf{F} & \textbf{Z} \\ \midrule

\multirow{16}{*}{Math}
    & Llama3.2-3B               & 0.0 & 11.83 & 8.74 & 9.24 & 6.34 & 0.0 & 19.18 \\
    & Llama3.2-3B-Instruct      & 33.33 & 44.98 & 30.79 & 36.51 & 32.39 & 0.0 & 42.2 \\
    & Qwen2.5-3B                & 66.67 & 33.35 & 37.67 & 36.22 & 43.66 & 0.0 & 49.33 \\
    & Qwen2.5-3B-Instruct       & 66.67 & 59.44 & 50.94 & 51.91 & 55.63 & 100.0 & 58.44 \\
    & Qwen3-4B                  & 66.67 & 59.94 & 48.42 & 53.52 & 44.37 & 100.0 & 59.09 \\
    & Qwen3-4B-Instruct         & 33.33 & 71.72 & 62.78 & 66.28 & 66.9 & 100.0 & 58.75 \\
    & Llama3.1-8B               & 33.33 & 28.88 & 20.38 & 21.85 & 21.13 & 0.0 & 35.12 \\
    & Llama3.1-8B-Instruct      & 33.33 & 48.71 & 36.34 & 41.94 & 30.99 & 0.0 & 50.71 \\
    & Qwen2.5-7B                & 66.67 & 58.55 & 46.51 & 50.44 & 45.07 & 0.0 & 58.4 \\
    & Qwen2.5-7B-Instruct       & 33.33 & 49.01 & 40.25 & 49.41 & 38.03 & 100.0 & 51.66 \\
    & Qwen3-8B                  & 66.67 & 65.95 & 55.09 & 59.09 & 59.15 & 100.0 & 62.29 \\
    & Qwen3-8B-Instruct         & 33.33 & 72.76 & 63.15 & 65.98 & 70.42 & 100.0 & 66.95 \\
    & Qwen2.5-14B               & 33.33 & 42.05 & 49.88 & 49.56 & 53.52 & 100.0 & 58.7 \\
    & Qwen2.5-14B-Instruct      & 33.33 & 49.85 & 38.62 & 42.23 & 39.44 & 100.0 & 65.96 \\
    & Qwen3-14B                 & 66.67 & 61.23 & 52.06 & 54.99 & 56.34 & 100.0 & 61.51 \\
    & Qwen3-14B-Instruct        & 33.33 & 60.29 & 46.55 & 48.24 & 50.0 & 100.0 & 67.78 \\ \bottomrule
\end{tabular}
}
\end{table*}

\begin{table*}[t]
\centering
\caption{Model Performance for Biology Across Academic Specializations Dimensions.}
\label{tab:results_field_biology}
\renewcommand{\arraystretch}{1.0}
\resizebox{\textwidth}{!}{
\begin{tabular}{llccccccccc}
\toprule
\textbf{Discipline} & \textbf{Model} & \textbf{A} & \textbf{B} & \textbf{C} & \textbf{D} & \textbf{E} & \textbf{F} & \textbf{G} & \textbf{H} & \textbf{Z} \\ \midrule

\multirow{16}{*}{Biology}
    & Llama3.2-3B               & 23.81 & 27.0 & 28.21 & 30.29 & 25.0 & 36.96 & 21.14 & 23.85 & 6.78 \\
    & Llama3.2-3B-Instruct      & 23.81 & 31.75 & 41.03 & 30.29 & 25.0 & 38.2 & 19.8 & 27.31 & 3.39 \\
    & Qwen2.5-3B                & 33.33 & 36.5 & 33.33 & 34.29 & 27.78 & 41.61 & 29.53 & 35.38 & 27.12 \\
    & Qwen2.5-3B-Instruct       & 26.98 & 37.0 & 46.15 & 38.29 & 33.33 & 39.44 & 28.86 & 34.23 & 18.64 \\
    & Qwen3-4B                  & 28.57 & 44.0 & 46.15 & 41.71 & 36.11 & 45.03 & 46.64 & 42.69 & 20.34 \\
    & Qwen3-4B-Instruct         & 36.51 & 44.25 & 33.33 & 41.14 & 25.0 & 43.79 & 37.58 & 42.69 & 10.17 \\
    & Llama3.1-8B               & 28.57 & 39.25 & 38.46 & 32.57 & 25.0 & 40.37 & 33.56 & 38.85 & 15.25 \\
    & Llama3.1-8B-Instruct      & 36.51 & 39.75 & 43.59 & 34.86 & 30.56 & 42.86 & 35.91 & 40.77 & 6.78 \\
    & Qwen2.5-7B                & 38.1 & 42.0 & 41.03 & 40.0 & 41.67 & 45.34 & 38.26 & 44.62 & 20.34 \\
    & Qwen2.5-7B-Instruct       & 31.75 & 29.5 & 35.9 & 25.14 & 22.22 & 34.16 & 22.48 & 27.69 & 1.69 \\
    & Qwen3-8B                  & 41.27 & 49.75 & 66.67 & 48.0 & 58.33 & 51.86 & 45.3 & 49.62 & 20.34 \\
    & Qwen3-8B-Instruct         & 44.44 & 47.75 & 53.85 & 54.29 & 38.89 & 51.86 & 48.66 & 50.0 & 30.51 \\
    & Qwen2.5-14B               & 41.27 & 40.5 & 48.72 & 47.43 & 41.67 & 47.52 & 39.6 & 46.15 & 18.64 \\
    & Qwen2.5-14B-Instruct      & 41.27 & 45.25 & 56.41 & 44.57 & 58.33 & 49.07 & 43.96 & 49.62 & 22.03 \\
    & Qwen3-14B                 & 50.79 & 55.5 & 48.72 & 53.71 & 41.67 & 50.93 & 52.35 & 53.08 & 15.25 \\
    & Qwen3-14B-Instruct        & 53.97 & 55.0 & 61.54 & 50.86 & 58.33 & 53.73 & 51.34 & 55.77 & 20.34 \\ \bottomrule
\end{tabular}
}
\end{table*}

\begin{table*}[t]
\centering
\caption{Model Performance for Physics Across Academic Specializations Dimensions.}
\label{tab:results_field_physics}
\renewcommand{\arraystretch}{1.0}
\resizebox{0.9\textwidth}{!}{
\begin{tabular}{llccccccccc}
\toprule
\textbf{Discipline} & \textbf{Model} & \textbf{A} & \textbf{B} & \textbf{C} & \textbf{D} & \textbf{E} & \textbf{F} & \textbf{G} & \textbf{Z} \\ \midrule

\multirow{16}{*}{Physics}
    & Llama3.2-3B               & 22.08 & 17.0 & 15.48 & 15.21 & 15.22 & 16.69 & 13.45 & 70.35 \\
    & Llama3.2-3B-Instruct      & 36.36 & 21.0 & 12.13 & 16.48 & 15.02 & 15.13 & 13.74 & 77.3 \\
    & Qwen2.5-3B                & 35.06 & 18.0 & 16.32 & 18.4 & 16.21 & 18.73 & 13.74 & 77.77 \\
    & Qwen2.5-3B-Instruct       & 28.57 & 22.58 & 9.92 & 15.27 & 12.65 & 17.11 & 11.44 & 77.41 \\
    & Qwen3-4B                  & 20.78 & 13.0 & 11.58 & 13.21 & 13.83 & 15.61 & 11.99 & 78.09 \\
    & Qwen3-4B-Instruct         & 55.84 & 28.0 & 24.83 & 32.7 & 34.19 & 32.17 & 27.49 & 79.97 \\
    & Llama3.1-8B               & 31.17 & 16.0 & 17.99 & 19.95 & 19.76 & 20.29 & 18.13 & 80.23 \\
    & Llama3.1-8B-Instruct      & 46.75 & 24.0 & 20.08 & 22.95 & 21.94 & 22.93 & 17.54 & 81.28 \\
    & Qwen2.5-7B                & 31.17 & 27.0 & 15.06 & 18.85 & 17.59 & 20.65 & 16.67 & 78.61 \\
    & Qwen2.5-7B-Instruct       & 28.57 & 28.0 & 10.46 & 16.03 & 13.64 & 18.85 & 11.7 & 78.97 \\
    & Qwen3-8B                  & 48.05 & 30.0 & 20.22 & 28.87 & 24.7 & 24.85 & 23.98 & 83.89 \\
    & Qwen3-8B-Instruct         & 58.44 & 43.0 & 26.08 & 34.61 & 33.0 & 33.01 & 31.29 & 85.25 \\
    & Qwen2.5-14B               & 32.47 & 33.0 & 14.37 & 20.31 & 20.55 & 22.69 & 20.76 & 86.72 \\
    & Qwen2.5-14B-Instruct      & 41.56 & 31.0 & 15.06 & 20.95 & 23.12 & 21.13 & 17.84 & 87.5 \\
    & Qwen3-14B                 & 51.95 & 39.0 & 28.45 & 36.61 & 33.99 & 32.41 & 32.46 & 82.64 \\
    & Qwen3-14B-Instruct        & 54.55 & 36.0 & 27.89 & 34.52 & 34.78 & 32.05 & 30.12 & 83.47 \\ \bottomrule
\end{tabular}
}
\end{table*}

\begin{table*}[t]
\centering
\caption{Model Performance for Chemistry Across Academic Specializations Dimensions.}
\label{tab:results_field_chemistry}
\renewcommand{\arraystretch}{1.0}
\resizebox{0.85\textwidth}{!}{
\begin{tabular}{llccccccccc}
\toprule
\textbf{Discipline} & \textbf{Model} & \textbf{A} & \textbf{B} & \textbf{C} & \textbf{D} & \textbf{E} & \textbf{F} & \textbf{Z} \\ \midrule

\multirow{16}{*}{Chemistry}
    & Llama3.2-3B               & 11.76 & 16.47 & 10.77 & 9.82 & 4.26 & 14.57 & 42.79 \\
    & Llama3.2-3B-Instruct      & 21.57 & 15.9 & 15.08 & 13.21 & 10.64 & 19.21 & 45.64 \\
    & Qwen2.5-3B                & 25.49 & 18.21 & 16.0 & 17.53 & 14.89 & 19.87 & 47.57 \\
    & Qwen2.5-3B-Instruct       & 20.59 & 16.47 & 21.54 & 14.65 & 18.09 & 17.88 & 43.39 \\
    & Qwen3-4B                  & 37.25 & 27.46 & 19.69 & 27.43 & 34.04 & 33.77 & 53.73 \\
    & Qwen3-4B-Instruct         & 30.39 & 29.48 & 28.92 & 26.42 & 27.66 & 28.48 & 54.01 \\
    & Llama3.1-8B               & 21.57 & 21.39 & 19.69 & 18.46 & 14.89 & 21.19 & 48.63 \\
    & Llama3.1-8B-Instruct      & 23.53 & 22.25 & 20.92 & 16.43 & 15.96 & 17.88 & 51.2 \\
    & Qwen2.5-7B                & 33.33 & 25.14 & 21.54 & 20.07 & 22.34 & 23.18 & 50.39 \\
    & Qwen2.5-7B-Instruct       & 33.33 & 24.57 & 22.77 & 19.73 & 18.09 & 28.48 & 50.98 \\
    & Qwen3-8B                  & 32.35 & 24.28 & 21.85 & 25.74 & 22.34 & 24.5 & 54.85 \\
    & Qwen3-8B-Instruct         & 29.41 & 26.01 & 30.46 & 25.66 & 21.28 & 25.83 & 53.02 \\
    & Qwen2.5-14B               & 28.43 & 21.39 & 20.92 & 17.53 & 20.21 & 19.87 & 52.04 \\
    & Qwen2.5-14B-Instruct      & 29.41 & 27.17 & 30.46 & 23.29 & 24.47 & 25.83 & 46.59 \\
    & Qwen3-14B                 & 40.2 & 32.95 & 34.77 & 32.18 & 30.85 & 27.81 & 58.26 \\
    & Qwen3-14B-Instruct        & 39.22 & 37.28 & 37.23 & 32.51 & 28.72 & 30.46 & 39.91 \\ \bottomrule
\end{tabular}
}
\end{table*}

\begin{table*}[t]
\centering
\caption{Model Performance for Math Across Bloom's Cognitive Dimensions.}
\label{tab:results_bloom_math}
\renewcommand{\arraystretch}{1.0}
\resizebox{\textwidth}{!}{
\begin{tabular}{llcccccc}
\toprule
\textbf{Discipline} & \textbf{Model} & \textbf{Remember} & \textbf{Understand} & \textbf{Apply} & \textbf{Analyze} & \textbf{Evaluate} & \textbf{Create} \\ \midrule

\multirow{16}{*}{Math} 
    & Llama3.2-3B            & 21.05 & 0.0 & 12.89 & 6.44 & 33.33 & 0.0 \\
    & Llama3.2-3B-Instruct   & 47.37 & 0.0 & 50.20 & 22.63 & 53.33 & 0.0 \\
    & Qwen2.5-3B             & 56.14 & 0.0 & 49.37 & 21.34 & 46.67 & 0.0 \\
    & Qwen2.5-3B-Instruct    & 71.93 & 0.0 & 68.98 & 38.63 & 60.00 & 0.0 \\
    & Qwen3-4B               & 70.18 & 0.0 & 67.40 & 38.08 & 60.00 & 25.00 \\
    & Qwen3-4B-Instruct      & 85.96 & 0.0 & 78.77 & 53.08 & 73.33 & 25.00 \\
    & Llama3.1-8B            & 50.88 & 0.0 & 33.82 & 12.27 & 40.00 & 0.0 \\
    & Llama3.1-8B-Instruct   & 66.67 & 0.0 & 55.01 & 26.68 & 60.00 & 0.0 \\
    & Qwen2.5-7B             & 64.91 & 0.0 & 65.79 & 36.04 & 66.67 & 0.0 \\
    & Qwen2.5-7B-Instruct    & 52.63 & 0.0 & 54.28 & 34.03 & 40.00 & 25.00 \\
    & Qwen3-8B               & 75.44 & 0.0 & 73.01 & 45.18 & 73.33 & 25.00 \\
    & Qwen3-8B-Instruct      & 80.70 & 0.0 & 79.40 & 53.81 & 80.00 & 25.00 \\
    & Qwen2.5-14B            & 77.19 & 0.0 & 59.29 & 32.74 & 40.00 & 0.0 \\
    & Qwen2.5-14B-Instruct   & 42.11 & 0.0 & 45.75 & 40.64 & 53.33 & 37.50 \\
    & Qwen3-14B              & 77.19 & 0.0 & 69.74 & 41.03 & 53.33 & 12.50 \\
    & Qwen3-14B-Instruct     & 78.95 & 0.0 & 63.51 & 39.47 & 60.00 & 0.0 \\   
    \bottomrule
\end{tabular}
}
\end{table*}

\begin{table*}[t]
\centering
\caption{Model Performance for Biology Across Bloom's Cognitive Dimensions.}
\label{tab:results_bloom_biology}
\renewcommand{\arraystretch}{1.0}
\resizebox{\textwidth}{!}{
\begin{tabular}{llcccccc}
\toprule
\textbf{Discipline} & \textbf{Model} & \textbf{Remember} & \textbf{Understand} & \textbf{Apply} & \textbf{Analyze} & \textbf{Evaluate} & \textbf{Create} \\ \midrule

\multirow{16}{*}{Biology}
    & Llama3.2-3B               & 17.38 & 47.22 & 8.91 & 33.74 & 66.67 & 0.0 \\
    & Llama3.2-3B-Instruct      & 15.73 & 44.44 & 13.86 & 38.0 & 61.11 & 0.0 \\
    & Qwen2.5-3B                & 26.16 & 63.89 & 16.83 & 41.26 & 72.22 & 100.0 \\
    & Qwen2.5-3B-Instruct       & 25.99 & 58.33 & 12.87 & 41.59 & 66.67 & 100.0 \\
    & Qwen3-4B                  & 31.95 & 63.89 & 23.76 & 50.78 & 66.67 & 0.0 \\
    & Qwen3-4B-Instruct         & 25.5 & 52.78 & 29.7 & 50.34 & 61.11 & 100.0 \\
    & Llama3.1-8B               & 26.66 & 66.67 & 20.79 & 42.49 & 61.11 & 0.0 \\
    & Llama3.1-8B-Instruct      & 26.16 & 55.56 & 17.82 & 46.75 & 66.67 & 100.0 \\
    & Qwen2.5-7B                & 28.81 & 66.67 & 25.74 & 49.78 & 66.67 & 100.0 \\
    & Qwen2.5-7B-Instruct       & 10.1 & 38.89 & 2.97 & 40.58 & 72.22 & 100.0 \\
    & Qwen3-8B                  & 35.76 & 69.44 & 27.72 & 57.62 & 83.33 & 100.0 \\
    & Qwen3-8B-Instruct         & 38.58 & 61.11 & 36.63 & 56.28 & 77.78 & 100.0 \\
    & Qwen2.5-14B               & 32.95 & 66.67 & 20.79 & 50.45 & 66.67 & 100.0 \\
    & Qwen2.5-14B-Instruct      & 36.42 & 55.56 & 29.7 & 53.25 & 72.22 & 100.0 \\
    & Qwen3-14B                 & 39.24 & 66.67 & 35.64 & 60.43 & 66.67 & 100.0 \\
    & Qwen3-14B-Instruct        & 41.56 & 72.22 & 31.68 & 61.43 & 72.22 & 100.0 \\  \bottomrule
\end{tabular}
}
\end{table*}

\begin{table*}[t]
\centering
\caption{Model Performance for Physics Across Bloom's Cognitive Dimensions.}
\label{tab:results_bloom_physics}
\renewcommand{\arraystretch}{1.0}
\resizebox{\textwidth}{!}{
\begin{tabular}{llcccccc}
\toprule
\textbf{Discipline} & \textbf{Model} & \textbf{Remember} & \textbf{Understand} & \textbf{Apply} & \textbf{Analyze} & \textbf{Evaluate} & \textbf{Create} \\ \midrule

\multirow{16}{*}{Physics}
    & Llama3.2-3B               & 27.86 & 43.33 & 11.13 & 20.36 & 25.0 & 0.0 \\
    & Llama3.2-3B-Instruct      & 32.14 & 56.67 & 9.38 & 21.35 & 31.25 & 0.0 \\
    & Qwen2.5-3B                & 23.57 & 63.33 & 12.45 & 23.46 & 12.5 & 0.0 \\
    & Qwen2.5-3B-Instruct       & 19.29 & 53.33 & 8.07 & 21.9 & 25.0 & 0.0 \\
    & Qwen3-4B                  & 31.43 & 10.0 & 11.38 & 14.81 & 18.75 & 0.0 \\
    & Qwen3-4B-Instruct         & 39.29 & 86.67 & 25.93 & 36.09 & 43.75 & 0.0 \\
    & Llama3.1-8B               & 30.71 & 80.0 & 15.28 & 23.86 & 18.75 & 0.0 \\
    & Llama3.1-8B-Instruct      & 37.86 & 73.33 & 15.77 & 28.88 & 31.25 & 0.0 \\
    & Qwen2.5-7B                & 32.86 & 66.67 & 13.38 & 23.53 & 31.25 & 0.0 \\
    & Qwen2.5-7B-Instruct       & 17.86 & 76.67 & 9.47 & 22.27 & 31.25 & 0.0 \\
    & Qwen3-8B                  & 42.14 & 83.33 & 19.09 & 31.99 & 43.75 & 0.0 \\
    & Qwen3-8B-Instruct         & 45.71 & 90.0 & 27.64 & 37.34 & 43.75 & 0.0 \\
    & Qwen2.5-14B               & 30.71 & 73.33 & 14.75 & 26.31 & 31.25 & 0.0 \\
    & Qwen2.5-14B-Instruct      & 32.86 & 80.0 & 13.38 & 28.49 & 37.5 & 0.0 \\
    & Qwen3-14B                 & 45.71 & 86.67 & 28.56 & 38.66 & 50.0 & 0.0 \\
    & Qwen3-14B-Instruct        & 52.86 & 83.33 & 26.27 & 38.93 & 56.25 & 0.0 \\  \bottomrule
\end{tabular}
}
\end{table*}

\begin{table*}[t]
\centering
\caption{Model Performance for Chemistry Across Bloom's Cognitive Dimensions.}
\label{tab:results_bloom_chemistry}
\renewcommand{\arraystretch}{1.0}
\resizebox{\textwidth}{!}{
\begin{tabular}{llcccccc}
\toprule
\textbf{Discipline} & \textbf{Model} & \textbf{Remember} & \textbf{Understand} & \textbf{Apply} & \textbf{Analyze} & \textbf{Evaluate} & \textbf{Create} \\ \midrule
\multirow{16}{*}{Chemistry}
    & Llama3.2-3B               & 16.36 & 38.46 & 6.68 & 13.91 & 20.0 & 0.0 \\
    & Llama3.2-3B-Instruct      & 15.15 & 53.85 & 8.37 & 19.36 & 30.0 & 0.0 \\
    & Qwen2.5-3B                & 23.64 & 61.54 & 14.66 & 18.8 & 60.0 & 0.0 \\
    & Qwen2.5-3B-Instruct       & 26.06 & 76.92 & 10.77 & 19.64 & 50.0 & 0.0 \\
    & Qwen3-4B                  & 30.3 & 69.23 & 24.93 & 28.29 & 40.0 & 0.0 \\
    & Qwen3-4B-Instruct         & 26.06 & 69.23 & 21.64 & 32.42 & 40.0 & 0.0 \\
    & Llama3.1-8B               & 24.85 & 46.15 & 13.36 & 22.56 & 60.0 & 0.0 \\
    & Llama3.1-8B-Instruct      & 26.06 & 46.15 & 13.86 & 21.05 & 30.0 & 0.0 \\
    & Qwen2.5-7B                & 38.18 & 69.23 & 14.96 & 25.09 & 50.0 & 0.0 \\
    & Qwen2.5-7B-Instruct       & 25.45 & 84.62 & 16.55 & 25.47 & 40.0 & 0.0 \\
    & Qwen3-8B                  & 38.18 & 61.54 & 20.84 & 25.85 & 50.0 & 0.0 \\
    & Qwen3-8B-Instruct         & 37.58 & 69.23 & 21.44 & 27.91 & 70.0 & 0.0 \\
    & Qwen2.5-14B               & 35.15 & 46.15 & 14.96 & 20.77 & 30.0 & 0.0 \\
    & Qwen2.5-14B-Instruct      & 42.42 & 69.23 & 18.44 & 29.04 & 40.0 & 0.0 \\
    & Qwen3-14B                 & 44.24 & 92.31 & 24.63 & 36.75 & 70.0 & 0.0 \\
    & Qwen3-14B-Instruct        & 49.09 & 92.31 & 26.62 & 37.22 & 60.0 & 0.0 \\  
    \bottomrule
\end{tabular}
}
\end{table*}

\end{document}